\documentclass[preprint,12pt]{elsarticle}

\usepackage{datetime}
\usepackage[dvipsnames,svgnames,x11names]{xcolor}

\usepackage{graphicx}
\usepackage{amssymb}
 \usepackage{amsthm}

\usepackage{lineno}

\usepackage{tikz}
\usepackage{mathtools}
\usepackage{subcaption}
\usepackage{afterpage}
\usepackage{hyphenat}

\usepackage[markup=underlined]{changes}
\definechangesauthor[color=NavyBlue]{CJM}

\usepackage[linesnumbered,ruled,vlined]{algorithm2e}
\SetKwInOut{KwInput}{input}
\SetKwInOut{KwOutput}{output}
\usepackage{booktabs}
\usepackage{multirow}
\usepackage{adjustbox}

\journal{Robotics and Autonomous Systems}

\begin{document}


\newcommand{\cNposes}{t}
\newcommand{\cNmappoints}{n}
\newcommand{\cvec}[1]{\mathbf{#1}}
\newcommand{\cMat}[1]{\mathbf{#1}}
\newcommand{\cInfo}[1]{\boldsymbol{\Lambda}_{#1}}
\newcommand{\cCov}[1]{\boldsymbol{\Sigma}_{#1}}
\newcommand{\cipq}{\cvec{q}}
\newcommand{\cipQ}{\cMat{Q}}
\newcommand{\cipx}{\cvec{s}}
\newcommand{\cipb}{\cvec{b}}
\newcommand{\cipN}{\cmappoints}
\newcommand{\cipk}{k}
\newcommand{\cipLambda}{\lambda}
\newcommand{\cipZeta}{\boldsymbol{\zeta}}
\newcommand{\cEntropy}[1]{H(#1)}
\newcommand{\cProb}[1]{P(#1)}
\newcommand{\cTranspose}[1]{#1^\text{T}}
\newcommand{\cSetS}{S}
\newcommand{\cSetV}{V}
\newcommand{\cSetR}{R}
\newcommand{\cmappointObs}[1]{o_{#1}}
\newcommand{\cmappointsObs}[1]{\cvec{o}}
\newcommand{\cmappoints}{\cvec{m}_{1:\cNmappoints}}
\newcommand{\cmappoint}[1]{\cvec{m}_{#1}}
\newcommand{\cposes}{\cvec{x}_{1:\cNposes}}
\newcommand{\cpose}[1]{\cvec{x}_{#1}}
\newcommand{\cobservations}{\cvec{z}_{1:\cNposes}}
\newcommand{\cobservation}[1]{\cvec{z}_{#1}}
\newcommand{\cpairwiseMin}[2]{\min(#1,#2)}
\newcommand{\cmatAtXY}[3]{}
\newcommand{\cmatAtX}[2]{}
\newcommand{\cmatAtY}[3]{}
\newcommand{\argmin}{\mathop{\mathrm{argmin}}}
\newcommand{\argmax}{\mathop{\mathrm{argmax}}}

\newcommand\inputpgf[2]{{
\let\pgfimageWithoutPath\pgfimage
\renewcommand{\pgfimage}[2][]{\pgfimageWithoutPath[##1]{#1/##2}}
\input{#1/#2}
}}

\newcommand{\cadd}[2]{\added[id=CJM,comment={#2}]{#1}}

\newcommand{\cftxt}[3]{
\raisebox{#1\height}{\begin{varwidth}{#2} \centering #3 \end{varwidth}}}
\newcommand{\cftxtleft}[3]{
\raisebox{#1\height}{\begin{varwidth}{#2} \raggedleft #3 \end{varwidth}}}


\begin{frontmatter}


\title{Map Point Selection for Visual SLAM}



\author[stel]{Christiaan J. M\"uller}
\author[stel]{Corné E. van Daalen}
\address[stel]{Department of Electrical and Electronic Engineering, Stellenbosch University, Stellenbosch, South Africa}

\begin{abstract}

Simultaneous localisation and mapping (SLAM) play a vital role in autonomous robotics. Robotic platforms are often resource-constrained, and this limitation motivates resource-efficient SLAM implementations. While sparse visual SLAM algorithms offer good accuracy for modest hardware requirements, even these more scalable sparse approaches face limitations when applied to large-scale and long-term scenarios. A contributing factor is that the point clouds resulting from SLAM are inefficient to use and contain significant redundancy. 

This paper proposes the use of subset selection algorithms to reduce the map produced by sparse visual SLAM algorithms. Information-theoretic techniques have been applied to simpler related problems before, but they do not scale if applied to the full visual SLAM problem. This paper proposes a number of novel information\hyp{}theoretic utility functions for map point selection and optimises these functions using greedy algorithms. The reduced maps are evaluated using practical data alongside an existing visual SLAM implementation (ORB-SLAM 2). Approximate selection techniques proposed in this paper achieve trajectory accuracy comparable to an offline baseline while being suitable for online use. These techniques enable the practical reduction of maps for visual SLAM with competitive trajectory accuracy.

Results also demonstrate that SLAM front-end performance can significantly impact the performance of map point selection. This shows the importance of testing map point selection with a front-end implementation. To exploit this, this paper proposes an approach that includes a model of the front-end in the utility function when additional information is available. This approach outperforms alternatives on applicable datasets and highlights future research directions.
\end{abstract}


\end{frontmatter}


\let\oldequation\equation
\let\oldendequation\endequation

\renewenvironment{equation}
  {\linenomathNonumbers\oldequation}
  {\oldendequation\endlinenomath}

\let\oldalign\align
\let\oldendalign\endalign
\renewenvironment{align}
{\linenomathNonumbers\oldalign}
{\oldendalign\endlinenomath}

\section{Introduction}\label{sec:intro}

Simultaneous localisation and mapping (SLAM) is the joint estimation of the pose of a robot and the state of its environment from sensor measurements. SLAM is essential for robustly addressing the autonomous navigation problem for mobile robots in unknown environments \cite{thrun2002probabilistic}. As shown by \citet{zaffar2018sensors}, cameras are a popular choice of sensor for SLAM due to their low cost and power consumption. These so-called visual SLAM algorithms can be categorised by map representation as either sparse or dense. While sparse algorithms produce less detailed maps than their dense counterparts, they require only modest hardware. Sparse visual SLAM have been demonstrated in applications of increasing scale over the past three decades (e.g. \cite{davison2007monoslam,klein2007parallel,mur2015orb}). 

Sparse SLAM algorithms abstract raw images into measurements of abstracted map points using visual feature techniques. These SLAM algorithms employ various heuristics to trade off the SLAM algorithm's accuracy and resource usage. Some of these heuristics include extracting a limited number of feature measurements per frame, using a subset of frames during SLAM estimation \cite{klein2007parallel} or pruning recently added map points with a limited number of observations \cite{mur2015orb}. Despite these approximations, \citet{cadena2016past} argues that the scalability of modern SLAM algorithms remains a challenge in large-scale environments and during long-term operation. In such scenarios, the map can grow unbounded, and consequently also the hardware requirements of the algorithms that use that map. This growth limits the applicability of SLAM for platforms with limited computational resources, memory, or communication bandwidth (in multi-robot scenarios). As a result, it is desirable to maintain an efficient representation of the environment to avoid exceeding these resource constraints. 

Existing approaches to limit resource usage are typically only capable of limiting the \emph{growth} of the map and are adjusted by experts in advance. This manual approach has a number of drawbacks: it introduces approximations even when surplus resources are available; it does not simplify the map when resources are limited; and the requirement of prior adjustment also limits the practical use when little information is known about the operational environment or an expert is not available to manually adjust the algorithm. To prevent these limitations, it is desirable to instead simplify the map in an automated fashion based on the hardware constraints during operation. This approach would introduce fewer approximations when surplus resources are available and only restrict resource usage when necessary.

This paper formulates the problem of selecting map points for visual SLAM as an optimisation problem. For this optimisation problem, it is necessary to define a utility function that represents the ``quality'' of a set of map points. Maximising the utility for a fixed number of points therefore selects the best set of points. The main focus of this paper is developing and evaluating different utility functions that can be efficiently optimised using existing optimisation algorithms. In the ideal case, a utility function should allow significant reductions to the map while minimally affecting SLAM performance and be computationally efficient to allow use in online applications.

Visual SLAM algorithms store both the measurements of map points used in estimation and the associated feature descriptors. To limit the size of the map when selecting map points, we assume that all information related to discarded map points is discarded. Only the measurement information and feature descriptors of selected map points are retained. 

A SLAM algorithm is typically divided into two components: 
the first component is the front\hyp{}end, which simplifies raw sensor measurements into 3D point measurements and performs data association. The second is the back-end, which estimates the robot pose and map from these processed measurements. Removing map points by performing map point selection will impact both the front\hyp{}end and back\hyp{}end performance. Better insight into these effects and good solutions to map point selection contributes to the development of selection approaches that consider a wider range of approximation techniques or hardware constraints. 

We identified two main approaches in the literature for map point selection in problems related to visual SLAM. The first approach is coverage\hyp{}based, where the problem of selecting map points is formulated in terms of variations of the set multi\hyp{}cover problem or maximum coverage problem. For this approach, map points are selected based on the number of map points visible in each camera. These coverage-based approaches have been applied to selecting map points for localisation (e.g. \cite{dymczyk2015gist,van2018efficient,cheng2016data}) or visual\hyp{}inertial navigation (e.g. \cite{lynen2015get,burki2016appearance,lynen2020large}). While these approaches could offer a computationally efficient means to select map points for SLAM, they do not consider the SLAM estimation problem.

An alternative approach for map point selection for SLAM is using information\hyp{}theoretic techniques. Information\hyp{}theoretic approaches are typically more expensive than the previously mentioned coverage\hyp{}based approaches, but explicitly optimise for estimation uncertainty. Information\hyp{}theoretic approaches have yet to be applied to the problem of selecting map points for the full visual SLAM problem, but were used for related problems. Previous approaches investigated information\hyp{}theoretic approaches for selecting a limited number of map point \emph{measurements} for only the last timestep during SLAM (e.g. \cite{kopitkov2017no,zhao2018good}). \citet{carlone2018attention} investigated selecting map points for visual\hyp{}inertial navigation, which is equivalent to SLAM in a limited window with an inertial measurement unit. These approaches could be adapted to selecting map points for the full SLAM trajectory, but require further modifications to be feasibly applied to the increased scale of selecting map points for SLAM. Additionally, selecting map points for SLAM also involve front-end performance, which is not considered by existing information\hyp{}theoretic approaches. 

The main contribution of this paper is the development and evaluation of a number of information\hyp{}theoretic approaches for map point selection for visual SLAM. This paper proposes several utility functions that can be optimised using existing efficient greedy techniques. The first utility function is a baseline that uses the entropy of the SLAM posterior distribution. This approach is similar to the one used by \citet{carlone2018attention} for visual\hyp{}inertial odometry. The formulation is adapted to the map point selection problem for SLAM. Different approximations and algorithmic improvements are made to improve scaling to be applicable to SLAM. Even with these improvements, this approach is not suitable for online use, but still serves as a performance baseline for more scalable approaches. This paper also proposes novel approximations of the baseline SLAM utility based on approximating it as either localisation or visual odometry problems. These approximations greatly improve the scalability compared to the baseline SLAM utility and reduce execution times from multiple hours to a few seconds for larger data sets. The software implementation of all methods developed in this paper are publicly available\footnote{https://github.com/ChristiaanM/MapSelect}. 

A further contribution is the development of an combined approach that augments the information\hyp{}theoretic utility function with a coverage\hyp{}based term based on loop\hyp{}closure information. This combined approach is still compatible with greedy optimisation algorithms. This additional term results in improved performance in relevant datasets. 

This paper is structured as follows: Relevant literature on existing approaches for related problems is discussed in Section \ref{sec:lit}. After that, Section \ref{sec:framework} provides a framework which introduces the map point selection framework used in this paper which includes greedy optimisation algorithms and the theory necessary for the information-theoretic approaches. The next two sections proposes utility functions to optimise within this framework: Section \ref{sec:traj} proposes information-theoretic utility functions, which includes both the baseline SLAM and approximate approaches, and is aimed at selecting map points for the back-end estimation portion of the SLAM implementation. 

The augmentation of information-theoretic utility functions with a coverage-based utility is presented in Section \ref{sec:loop}, which aims to improve the information-theoretic utilities by assigning higher utilities for map points relevant for identifying loop closures. To evaluate the proposed techniques, we first discuss some of the challenges of using an existing SLAM implementation, then motivate software changes and a testing framework in Section \ref{sec:setup}. This framework is then used to provide experimental results presented in Section \ref{sec:experiment}. Lastly, we provide some concluding remarks and present avenues for future research in Section \ref{sec:conclusion}.
 
\section{Related Work}\label{sec:lit}
 This section reviews existing work related to map point selection for sparse visual SLAM. It provides an overview of sparse visual SLAM algorithms, coverage\hyp{}based map point selection approaches developed for localisation and information\hyp{}theoretic selection approaches.
 
    \subsection{Visual SLAM Algorithms}
        In early approaches to visual SLAM, the use of recursive estimators was popular and \citet{davison2007monoslam} proposed the first monocular SLAM algorithm using an extended Kalman filter (EKF). \citet{thrun2002probabilistic} provides a good overview of classical approaches to SLAM. Due to the poor asymptotic complexity, these approaches have mainly been superseded by graph\hyp{}based SLAM approaches. 
        
        \citet{klein2007parallel} identified that not all camera images are useful for estimating the map and introduced keyframe\hyp{}based SLAM techniques with the parallel tracking and mapping (PTAM) algorithm. This approach approximates the localisation of the camera in real time by visual tracking, ignoring map uncertainty. The SLAM estimation is done in parallel at a slower frequency, using bundle adjustment with a spatially distributed subset of the camera frames called keyframes. Bundle adjustment jointly estimates the full trajectory of camera poses instead of only the current robot pose and is the gold standard in offline reconstruction. \citet{strasdat2012visual} showed that bundle adjustment results in more accurate trajectory estimates for visual SLAM than filtering at the same computational budget.
        
        \citet{strasdat2011double} proposed performing bundle adjustment in a local window while using approximate pose-graph optimisation in a larger outer window. Pose-graph optimisation approximates the marginal distribution over camera poses with approximate heuristic factors based on the co-visibility or visual overlap between different camera poses. 
        
        \citet{mur2015orb} then developed the ORB-SLAM algorithm, which built on these initial approaches. ORB-SLAM further separated the problem into different specialised sub-components running in parallel. Loop closures were detected using the visual large-scale localisation technique proposed by \citet{galvez2012bags}. SLAM estimation was based on keyframe approaches, while large-scale loop closures were optimised in parallel using an initial pose-graph heuristic followed by bundle adjustment. \citet{mur2017orb} extended the baseline ORB-SLAM algorithm in ORB-SLAM 2 to support stereo cameras. \citet{ORBSLAM3} developed ORB-SLAM 3, which added support for inertial measurement units, revised the loop\hyp{}closure algorithm and added improved support for multi-session mapping. 
        
         As an alternative to the aforementioned feature\hyp{}based (or indirect) methods, direct methods have also been proposed to estimate the map by optimising directly on image intensities. \citet{engel2014lsd} proposed LSD-SLAM, a direct method for doing visual SLAM. However, this approach still required feature\hyp{}based techniques for loop\hyp{}closure detection and was shown to be less accurate than feature\hyp{}based methods by \citet{mur2017orb}. While less robust than direct methods to the lack of texture or motion blur in images, \citet{ORBSLAM3} shows ORB-SLAM 2 and 3 remain state of the art in terms of trajectory accuracy. We choose to use ORB-SLAM 2 in this paper as a representative SLAM implementation to evaluate the effects of map point selection algorithms. This choice is discussed in greater detail in Section \ref{sec:setup}. 
    \subsection{Coverage-based Map Point Selection for Localisation}
        This section considers coverage-based map point selection approaches used for feature\hyp{}based maps. Various works have investigated selecting map points for the large-scale maps produced by sparse feature\hyp{}based structure-from-motion algorithms. These structure-from-motion algorithms can be seen as an offline version of visual SLAM, without ordered camera images or real-time processing constraints. The resulting reduced maps are then used for localisation instead of SLAM.
        
       \citet{li2010location} identified that many map points are redundant for localisation and proposed making a sub-selection of the map to improve localisation efficiency. They proposed to model this problem as a set multi-cover problem, which involves selecting the smallest set of map points where each camera view retains at least a specified number of map points. The set multi-cover problem is a generalisation of the set cover problem. While the set multi-cover problem is known to be NP-hard, it can be approximated using greedy algorithms in polynomial time.
       
       \citet{cheng2016data,cao2014minimal} and \citet{camposeco2019hybrid} investigated adaptations to improve the baseline set multi-cover approach for large-scale localisation. These approaches emphasise more unique descriptors, higher observation counts, or well-distributed points in the image plane. Emphasis on these properties improved localisation performance over the baseline set multi-cover approach. While often without strong optimisation guarantees, these heuristics resulted in scalable approaches for map point selection for large-scale localisation. 
        
       As an alternative to heuristics, integer programming techniques for solving map coverage style problems have also been proposed by \citet{van2018efficient,soo20133d} and \citet{dymczyk2015keep}. Integer programming techniques find the optimal solution at the expense of poor worst-case computational complexity. \citet{soo20133d} proposed both linear and quadratic integer programs for reducing maps for localisation. \citet{dymczyk2015keep} adapted the formulation by \citet{soo20133d} to using the set coverage as a soft constraint, while instead limiting the total number of map points to a certain maximum value. This approach proved remarkably scalable in practice. \citet{lynen2020large} recently demonstrated this approach on city-scale problems by dividing a city into blocks of 150-metres and selecting map points for each block independently. Additionally, \citet{van2018efficient} proposed a feature coding scheme and revised the formulation of \citet{dymczyk2015keep} to account for different storage costs associated with compressed feature descriptors.
        
        A few approaches also included visual\hyp{}inertial tracking to allow high-frequency updates when localising in reduced maps after map point selection. Greedy heuristics have been used by \citet{lynen2015get} for this problem, while \citet{schneider2018maplab} and \citet{lynen2020large} investigated integer programming techniques. \citet{schneider2018maplab} developed Maplab and the associated robust visual\hyp{}inertial odometry with localisation integration (ROVIOLI) algorithm. This algorithm augments the baseline direct vision odometry technique with offline tools for bundle adjustment and uses the resulting map to assist localisation. The localisation and map-building approaches are feature\hyp{}based and use the integer program proposed by \citet{dymczyk2015keep} to reduce the offline map. 
        
        This overview of existing work shows that coverage\hyp{}based models can simplify maps efficiently for large-scale localisation but have not yet been applied to SLAM. Visual SLAM implementations employ similar localisation algorithms to perform large\hyp{}scale loop closures (e.g. \cite{mur2015orb,engel2014lsd}). Due to the success of these coverage\hyp{}based algorithms for localisation, we expect this approach would also do well for the localisation subsystem of a SLAM implementation. However, these coverage\hyp{}based approaches will not necessarily allow for accurate SLAM trajectory estimation.

    \subsection{Information-theoretic Selection Approaches}
        We now consider existing information\hyp{}theoretic selection approaches related to the selection of map points for SLAM. These approaches address different selection problems for SLAM, visual odometry or localisation with respect to a statistical measure (typically entropy or information gain). 
        
        Early approaches to SLAM used Kalman filtering to estimate both the map and current robot position. A Kalman filter requires the maintenance of a dense covariance matrix of the robot pose and landmarks. For Kalman filters it is essential to limit the number of landmarks to ensure real-time operation. Early approaches by \citet{dissanayake2000computationally} and \citet{zhang2005entropy} proposed methods to prioritise landmarks by creating new landmarks based on information gain while discarding landmarks no longer being observed.
        
        In later work, \citet{choudhary2015information} investigated the selection of landmarks for 2D graph SLAM. They investigated a heuristic formulation to trade off the number of map points in the map against information gain. While this approach allowed for the reduction of map size in practice, the greedy heuristic does not provide theoretical guarantees or limit the number of map points in the map. 

        \citet{kopitkov2017no} showed the similarity between different decision\hyp{}making problems under uncertainty and proposed a general algorithm for exploiting sparsity. The assumption of known future observation results in the equivalence between different optimisation problems, such as the selection map points or actions a robot must perform to maximise information gain. They applied this general algorithm to the problem of selecting map point measurements at the current timestep for SLAM.

        \citet{zhao2018good} limited the tracking time for ORB-SLAM by selecting map point measurements based on modelling the localisation problem of the current camera frame or pose. At each timestep, only a limited number of feature measurements were selected to be used for estimation. To improve the scalability of the selection algorithm, they used the stochastic greedy algorithm by \citet{mirzasoleiman2014lazier}. The computational complexity was further improved by using Cholesky updates to avoid recomputing the entropy from scratch at each step of the greedy algorithm. This feature selection approach was tested using the output of the tracking module of the ORB-SLAM 2 algorithm that performs localisation. Therefore, this approach did not test the effect of selection on the SLAM estimation or limit the total number of map points in the map. \citet{zhao2020good} proposed limiting computation time for visual SLAM estimation by selecting the camera poses to include in SLAM estimation. Importantly, they use similar techniques to their earlier work to speed up computation.

        \citet{carlone2018attention} proposed an information\hyp{}theoretic technique for selecting map points during visual\hyp{}inertial odometry to limit computation time. Visual\hyp{}inertial odometry is equivalent to the local SLAM estimation problem that only considers the most recent set of poses and the associated map points. Visual\hyp{}inertial odometry involves the use of an inertial measurement unit (IMU) in addition to a camera. This approach made two important assumptions appropriate for this setting: the orientation uncertainty is negligible due to the IMU, and the future states are known due to access to the robot's commands. This approach used greedy algorithms to optimise the proposed utility function with guarantees.
        
        Selecting map points for visual SLAM given a specified budget remains unexplored. The work by \citet{carlone2018attention} is the closest to the problem investigated by this paper, but selecting map points for SLAM has two crucial differences from selecting map points for visual odometry. One of the differences is that SLAM presents a multiple order of magnitude increase in the number of map points and poses. \citet{carlone2018attention} selected ten map points from a hundred, while SLAM maps can easily contain hundreds of thousands of map points. Different approximations are necessary to select map points in problems of this scale.
                
        Another difference between selecting map points for SLAM instead of visual odometry is that SLAM implementations involve multiple interacting sub-modules to address the SLAM problem. These modules focus on subproblems such as updating the local SLAM estimation, detecting loop closures and large-scale estimation after loop closures. Selecting map points for a visual SLAM implementation will influence these respective sub-modules. 
                
        The baseline SLAM approach in this paper adapts these existing information-theoretic selection approaches to the map point selection problem for SLAM. The approach is similar to that used by \citet{carlone2018attention} for visual\hyp{}inertial odometry without IMU information. This paper also does not assume known orientation or model future poses. Implementation optimisations from \citet{zhao2018good} and \citet{zhao2020good} in related problems are used to improve the asymptotic complexity of the proposed approach. 
    
\section{Map Point Selection for SLAM}\label{sec:framework}
This section defines the map point selection problem for SLAM and provides the necessary techniques to develop and optimise the utility functions in this paper. Subsection \ref{subsec:framework} covers the general framework used by this paper. Relevant techniques to optimise utility functions are covered in Section \ref{subsec:submod}. Lastly, concepts related to the SLAM problem necessary to develop the information-theoretic utility functions for this paper are covered in Subsection \ref{subsec:vslam}.

\subsection{Framework for Map Point Selection}\label{subsec:framework}  
    
    When formulating the problem of selecting map points, it is necessary to describe the utility of a set of points, which quantifies the value or quality of the points. Before we consider specific utility functions, we first describe the problem in terms of a general utility function. For the set of available map points, $V$, and a selected subset, $S$, there is a function $f(S)$ that models the utility of $S$. 
    We assume that the map's relevant hardware constraints for an application (whether it is the limited memory, storage space or communication budget) can be translated to a limited budget of map points $k$.  
    During map point selection, we wish to find the subset of map points that maximises the utility while satisfying the budget constraints, or 
       \begin{equation}\label{eq:mapselect}
         \operatorname*{argmax}\limits_{S \subseteq V, |S|\leq k } f(S). 
        \end{equation}
     Problems of this form are difficult in general, especially given stringent requirements on computational requirements for use in online applications. If we assume that $f(S)$ is monotone, submodular and normalised, greedy algorithms can approximately maximise (\ref{eq:mapselect}), with strong guarantees in polynomial time. These properties and relevant algorithms are discussed in greater detail in Subsection \ref{subsec:submod}. Due to the efficiency and guarantees of these algorithms, our approach is instead to suitable utility functions for the map point selection problem for SLAM. 
    
     Online visual SLAM implementations consists of different sub-components that approximate the SLAM problem and employ various heuristics. This includes for example real-time localisation, online SLAM estimation and loop-closure detection. Selecting map points will impact all sub-components that use the map. We suspect that to accurately describe the utility of a set of map points for a SLAM implementation, it would be beneficial to consider the impact of the selection on all these sub-components. However, such a hypothetical utility would likely be both challenging to obtain and efficiently optimise. This paper instead first proposes general information-theoretic models using a utility function based on back-end estimation problem that can be augmented to better capture the utility of the overall problem.

    \begin{figure}[h]
        \begin{center}
            \makebox[0.9\textwidth]{\includegraphics[width=0.9\textwidth]{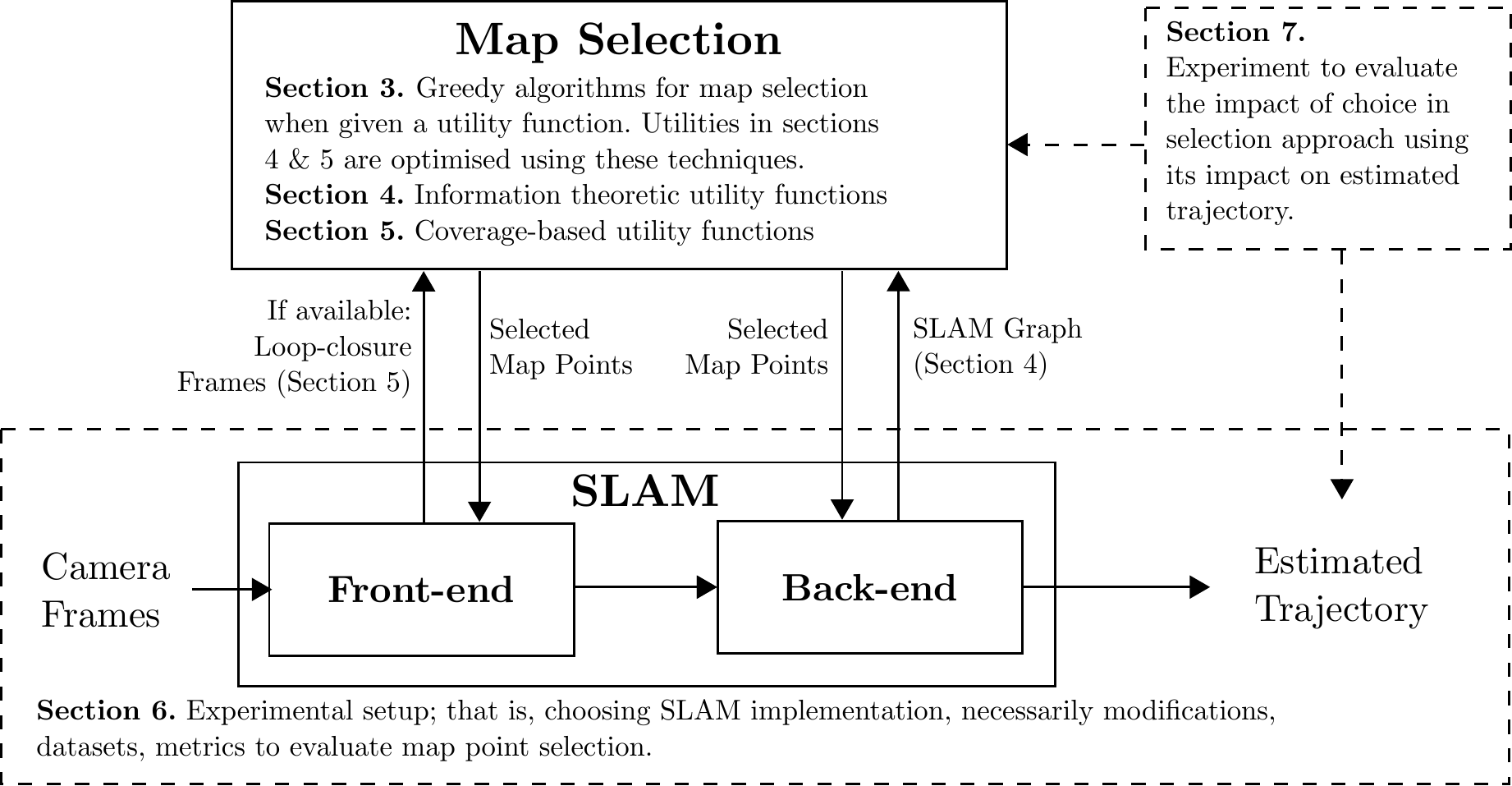}}
    \end{center}
        \caption{ 
           Overview of map point selection for SLAM and how it relates to relevant sections of the paper. Section 3 proposes a framework for formulating map point selection problems given a utility function. Sections 4 and 5 propose choices for this utility function. Section 6 considers the problem of evaluating map point selection approaches using a SLAM implementation. Different components of the SLAM implementation with map point selection are shown in solid outlines. We use solid arrows to indicate the flow of information between these components. Dashed boxes and arrows are used to indicate the relevance of sections to different components or emphasise the relationship of sections to items in the diagram. 
        }
        \label{fig:overview}
    \end{figure}

     Visual SLAM algorithms create new map points whenever keyframes are inserted. When map point selection is integrated alongside a SLAM algorithm, an optimisation algorithm is used at regular intervals to reduce the map to a specified budget. While selecting map points at every keyframe would be ideal, selection approaches are typically too computationally expensive to be performed at typical frame rates. Instead, we recommend selecting map points in parallel to the SLAM implementation if possible. This approach allows online map point selection to be performed at a lower frequency. When formulating the map point selection problem, information is used from the SLAM algorithm to formulate the optimisation problem. After selecting map points, the result is used to remove map points from the map used by the SLAM implementation.

    Figure 1 provides an overview of how the sections of this paper address the problem of map point selection for SLAM. This section (Section \ref{sec:framework}) covers the necessary optimisation algorithms and formulates the problem of selecting a subset of map points, a utility function and a budget constraint. The selected map impacts both the front-end and back-end of a SLAM implementation. Subsequently, Section \ref{sec:traj} proposes utility functions that approximate the utility of map points for SLAM estimation. These functions are not specific to any dataset or SLAM. Section \ref{sec:loop} proposes augmenting the utility function with an additional term to improve loop-closure detection performance by improving the model of the map point utility for a SLAM implementation. Section \ref{sec:setup} present as experimental setup for evaluating different map point selection approaches using a SLAM implementation. Lastly, Section \ref{sec:experiment} presents the experimental results from this experiment.
    
    \subsection{Submodular Maximisation}\label{subsec:submod}
        The utility functions in this paper that describe the utility of a set of map points belong to a class of set functions known as submodular functions. Submodular set functions have the important property that they can be approximately maximised using greedy algorithms for a variety of constraints. This subsection briefly describes submodular functions and greedy optimisation algorithms.
        
        For a discrete set function defined over set $V$ with 
        \begin{equation}
            f := 2^V \rightarrow \mathbb{R},\nonumber
        \end{equation}
        we define the marginal gain of element $f(m_i|S)$ as 
        \begin{equation}\label{eq:marggain}
           f(m_i|S) := f(S \cup \{m_i\}) - f(S),
        \end{equation}
        where $m_i$ is an element in the set. The function $f$ is submodular if, for any sets $A$ and $B$ with $A \subseteq B \subseteq V$, the marginal gain has the property of diminishing returns, or 
        \begin{equation}\label{eq:submod}
            f(m_i|A) \geq f(m_i|B).
        \end{equation}
        The set function $f(S)$ is monotone if, for any sets $A$ and $B$ with $A \subseteq B$, it has the property that
        \begin{equation}\label{eq:monotone}
            f(A) \leq f(B). 
        \end{equation}
        Lastly, the set function is normalised if, for the empty set $\emptyset = \{\}$, 
        \begin{equation}\label{eq:norm}
            f(\emptyset) = 0.
        \end{equation}
         Submodularity is closed under non-negative linear combinations. For two submodular functions, $f_{1}(S)$ and $f_{2}(S)$, and non-negative weights $w_{1}$ and $w_{2}$, the weighted sum $f_{3}(S)$, or 
        \begin{equation}\label{eq:sumsubmod}
            f_{3}(S) = w_{1}f_{1}(S)+w_{2}f_{2}(S),
        \end{equation}
        is submodular. 

       In selection problems, we are often interested in selecting some optimal subset $S^*$ of at most $k$ items that maximises a utility $f(S)$, or
       \begin{equation}
         S^*= \operatorname*{argmax}\limits_{S \subseteq V, |S|\leq k } f(S) \nonumber
        \end{equation}
        This problem is difficult in general and NP-hard for many cases of submodular functions \cite{nemhauser1978best}. Fortunately, greedy algorithms often provide strong guarantees for these classes of functions. If $f(S)$ is a monotone, normalised and submodular function subjected to cardinality constraint $k$, the classic greedy algorithm returns a $1-e^{-1}$ approximation \cite{Nemhauser1978}; that is, the obtained utility by the greedy solution $S_g$ is at least within a constant factor of the function value obtained by the intractable optimal set, or
        \begin{equation}
            f(S_g) \geq \left(1-e^{-1} \right) f(S^*). \nonumber
        \end{equation}
        
        \begin{algorithm}[H] \label{algo:greedymax}
            \KwInput{$f:2^V \xrightarrow{} \mathbb{R_+} \;\text{and}\; \cipk$ }
            \KwOutput{  $\cSetS$}
            $S \gets \emptyset$ \\
           \For{$i\gets 1$ \KwTo $k$}
           {
                $m_i \gets \argmax\limits_{m_i \in \cSetV \setminus \cSetS  }  f(m_i|\cSetS) $\\
                $\cSetS  \gets \cSetS \cup \{ m_i \}$ 
            }
            \caption{Classic greedy algorithm for monotone cardinality constrained submodular maximisation}
        \end{algorithm}
        
        As shown in Algorithm \ref{algo:greedymax}, the classic greedy algorithm requires $n \times k$ function evaluations of the marginal gain, or $f(m_i|S)$, where $n$ is the number of items in $V$ and $k$ is the budget or the number of selected items. If the budget grows proportional to $n$, the greedy algorithm requires $O(n^2)$ function evaluations. In practice, this can be reduced by maintaining the upper bound of the marginal gain for each remaining item in the set. This modification by \citet{minoux1978accelerated} yields the lazy greedy algorithm. While the lazy greedy algorithm shares the same worst\hyp{}case complexity as the classic greedy algorithm, which always evaluates all remaining elements, it is often significantly faster in practice.

        \begin{algorithm}[H] \label{algo:stochgreedy}
            \KwInput{$f:2^V \xrightarrow{} \mathbb{R_+}\;\text{,}\;\cipk\;\text{and}\;\epsilon$ }
            \KwOutput{  $\cSetS$}
            $r \gets \frac{n}{k} \log(\frac{1}{\epsilon})$ \\
            $S \gets \emptyset$ \\
            \For{$i\gets 1$ \KwTo $k$}
            {
                $\cSetR \gets$ Select random subset of $r$ elements from $V\setminus  \cSetS $ \\
                $m_i \gets  \argmax\limits_{m_i \in \cSetR  }   f(m_i|\cSetS) $\\
                $\cSetS  \gets \cSetS \cup \{ m_i \}$ \\
            }
            \caption{The stochastic greedy algorithm}
        \end{algorithm}
        
        An efficient stochastic variation of the greedy algorithm was proposed by \citet{mirzasoleiman2014lazier} and is shown in Algorithm \ref{algo:stochgreedy}. The stochastic greedy algorithm requires only $O(n\log(\frac{1}{\epsilon}))$ function evaluations, while having a weaker approximation guarantee of $1-e^{-1}-\epsilon$ and only on the expected function value of the utility, or
        \begin{equation}
            E[f(S_g)] \geq  \left(1-e^{-1} -\epsilon \right) f(S^*). 
        \end{equation}

        For many set functions, the complexity of evaluating the marginal gain $f(m_i|\cSetS)$ is a function of the number of elements in the set $n$. The resulting worst-case computational complexity for optimising utilities with the lazy greedy and stochastic greedy algorithms are $O(n^2\times\text{g}(n))$ and $O(n\times\text{g}(n))$, respectively, where $\text{g}(n)$ is the computational complexity of evaluating the marginal gain.

        Greedy algorithms provide strong guarantees for approximately maximising submodular utility functions and only require polynomial time. These greedy algorithms allow for the development of computationally efficient approaches to map point selection. Additionally, utility functions for which the marginal gain can be calculated efficiently will result in more scalable approaches.

        We primarily evaluate different map point selection approaches using the lazy greedy algorithm. Stochastic greedy incurs a performance penalty compared to lazy greedy in expectation. However, stochastic greedy is an important alternative when it is necessary to speed up selection for cases where it is not feasible to use lazy greedy.  
        
 \subsection{The Visual SLAM Problem}\label{subsec:vslam}
    One class of utility functions -- specifically the information-theoretic utility functions -- rely on a model of the SLAM inference performed in the back-end. The goal of SLAM is to jointly estimate the robot states and the map from a set of sensor measurements of the environment. Practical sensor measurements present various challenges that complicate a unified approach to solving the problem. Sensors such as cameras can produce very high-dimensional measurements that are computationally expensive to process and difficult to model robustly. Consequently, practical SLAM algorithms can be divided into two parts: the front-end and the back-end. The front-end prepossesses sensor measurements into a more convenient form using feature extraction and draws correspondence. The back-end then performs the joint estimation of poses and map points using these processed measurements.

    In sparse visual SLAM algorithms, visual-feature descriptors are used to describe the local image texture of selected pixel locations. These descriptors are designed to be viewpoint invariant and allow for drawing correspondence between abstracted map point observations between different time steps. This significantly simplifies the back-end estimation problem. The back-end formulates the SLAM problem as the joint estimation of the robot poses, or $\cposes$, a set of uniquely identifiable positions of map points, $\cmappoints$, and the respective pixel locations, $\cobservations$, where those points were observed, or 
    \begin{equation}\label{eq:slam}
        \cProb{\cposes,\cmappoints|\cobservations}.
    \end{equation}
     Note that this paper uses $n$ for the total number of map points and $t$ for the total number of estimated robot poses. Ranges, such as $1{:}n$, indicate the range of indices from $1$ to $n$. A further convention is that this paper generally uses $i$ as an index associated with map points and $j$ as an index associated with robot poses.

    Using the common assumptions that map point observations are conditionally independent and by applying Bayes' rule, the posterior distribution over the poses and map points can be factorised as 
        \begin{equation} \label{eq:slam-factored}
            \cProb{\cposes,\cmappoints|\cobservations} = \eta \cProb{\cposes} \prod\limits_{j=1}^{t} \prod\limits_{i=1}^{n} \cProb{\cobservation{j}^{i}|\cpose{i}\cmappoint{j}}.
        \end{equation}
    
     Maximum a posterior (MAP) estimation is commonly used in sparse visual SLAM. The MAP estimate is found by maximising of the posterior distribution, which is equivalent to minimising the negative log posterior. If we assume Gaussian noise, the SLAM estimation problem becomes a non-linear least squares problem. To simplify the notation, we introduce $\cvec{y}$, which is the combined vector of both $\cposes$ and $\cmappoints$. The least-squares problem is solved by successive approximations of the linear estimation problem around the current estimate, or   
    \begin{equation}\label{eq:linlogpost}
         l(\cvec{y}) = -\log \cProb{\cvec{y}|\cobservations}
                      \approx  \frac{1}{2}\cTranspose{\cvec{y}}\cInfo{\cvec{y}}\cvec{y} - \cTranspose{\cvec{b}}\cvec{y} + \text{const.}
    \end{equation}
    
    For a more in-depth discussion on least-squares estimators and SLAM implementations, we direct the interested reader to the tutorial by \citet{grisetti2010tutorial} and toolboxes developed for SLAM estimation (e.g. g\textsuperscript{2}o \cite{grisetti2011g2o}, GTSAM \cite{dellaert2012factor} and SLAM++ \cite{ila2017slam++}).
        
        For this paper, it is important to note that the information matrix of the linear approximation, or $\cInfo{\cvec{y}}$ in (\ref{eq:linlogpost}), can be used to approximate relevant information\hyp{}theoretic quantities of the non-linear posterior distribution. The approximate information matrix $\cInfo{\cvec{y}}$ can be calculated directly from all the quadratic terms from the linear approximation of the posterior in (\ref{eq:slam-factored}) as 
        \begin{equation}\label{eq:slam-sum}
            \cInfo{\cvec{y}} = \sum\limits_{j=1}^{t}\sum\limits_{i=1}^{n} \cTranspose{\cMat{J}_{ij}}\cMat{\Omega}_{ij}\cMat{J}_{ij},
        \end{equation}
        where $\cMat{J}_{ij}$ is the Jacobian of the measurement function of map point $i$ from pose $j$ and we assumed Gaussian measurement noise with an information matrix (or inverse covariance matrix) of $\cMat{\Omega}_{ij}$. This approach is used in Section \ref{sec:traj} for deriving information-theoretic utility functions.

       For a general Gaussian distribution over random variables $\cvec{y}$, the entropy is a function of the information matrix $\cInfo{y}$ and the dimension $d$ of the matrix, or
        \begin{equation}\label{eq:ent-gauss}
            \cEntropy{\cvec{y}} = \frac{d}{2} + \frac{d}{2}\log(2\pi) - \frac{1}{2} \log(|\cInfo{\cvec{y}}|).
        \end{equation}
        Similarly, the information gain for a general Gaussian distribution is given in terms of the difference between the posterior information matrix $\cInfo{\cvec{y}|\cvec{z}}$ and prior information matrix $\cInfo{y}$, or 
        \begin{equation}\label{eq:info-gain-gaus}
            IG(\cvec{y},\cvec{z}) = \frac{1}{2}\log(|\cInfo{\cvec{y}|\cvec{z}}|) - \frac{1}{2}\log(|\cInfo{\cvec{y}}|).
        \end{equation}
            
        To evaluate the information gain in (\ref{eq:info-gain-gaus}) as a function of a selected set of map points, we need to calculate the prior and posterior information matrices as a function of the selected map points. These information matrices can be calculated directly from the linear approximations of the respective distributions as shown in (\ref{eq:slam-sum}) and is used for the information-theoretic utility functions proposed by this paper.

\section{Information-theoretic Utility Functions for Map Point Selection}\label{sec:traj}
   
    This section proposes information\hyp{}theoretic utility functions for map point selection using the framework from the previous section. This section uses the information\hyp{}theoretic concept of information gain as a utility for SLAM\hyp{}based selection problems and is aimed at modelling utility for the back-end of the SLAM implementation. Subsection \ref{subsec:entropy} proposes a model based on the SLAM posterior. To improve the scalability of this baseline approach, approximate approaches based on localisation or visual odometry are proposed in Subsections \ref{subsec:local} and \ref{subsec:odom}, respectively.

    \subsection{Utility based on SLAM Entropy}\label{subsec:entropy}
        The performance of SLAM algorithms is often evaluated by the accuracy of the estimated robot trajectories. Therefore, a good choice for a utility function would be to model a point's contribution to the estimation of the robot's trajectory. This requires the consideration of potential future observations of those map points and the resulting potential beliefs. An approximation of this intractable full model is to use the SLAM model at the current time step and select map points to only minimise uncertainty over the past trajectory. 
        
         This paper uses a formulation similar to \citet{kopitkov2017no} and \citet{carlone2018attention} to maximise the information gain over the posterior distribution of the current robot poses. This approach is adapted to select a limited subset of map points from the full map instead of only selecting map points for the last pose or the most recent window of poses. We also do not predict nearby future poses using inertial\hyp{}measurement data as used by \citet{kopitkov2017no}.
        
        We define the utility of a selected set of map points $S$ as the information gained in the posterior distribution with respect to the prior distribution over robot poses. For Gaussian distributions, the information gain simplifies to the difference between the log\hyp{}determinant of the posterior information matrix, or $\cInfo{\cposes}^S$, and a constant term based on the prior distribution $c_{\emptyset}$, or 
    
        \begin{equation}\label{eq:slam-def}
             f_{\text{SLAM}}(S) = - \frac{1}{2} c_{\emptyset} + \frac{1}{2} \log(|\cInfo{\cposes}^S|).
        \end{equation}

        To calculate the information matrix $\cInfo{\cposes}^S$, we first manipulate the posterior distribution and then show how to calculate the information matrices directly from the posterior distribution. For the selected set of map points, $S$, we define $\cmappoint{S}$ as the map point positions corresponding to set $S$ and $\cobservations^S$ as the subset of observations corresponding to those map points. The posterior distribution over the robot poses, using only measurements of the selected set of map points $S$, is given by the marginal of the joint distribution over poses and map points, or $\cProb{\cposes|\cobservations^S}$. We manipulate the posterior into the factors associated with each map point while noting the associated information matrices, or 
        \begin{equation}
            \underbrace{\cProb{\cposes|\cobservations^S}}_{\cInfo{\cposes}^S}
                =  \eta \underbrace{\cProb{\cposes}}_{\epsilon\cMat{I}} \prod_{i \in S} 
                \underbrace{
                    \int \overbrace{\cProb{\cobservations^{i}|\cposes,\cmappoint{i}}}^{\cInfo{\cposes,\cmappoint{i}}}\text{d}\cmappoint{i}.
                }_{\cInfo{\cposes}^i} \label{eq:ent-sum-probs}
        \end{equation} 
     
         We assume an independent and identically distributed Gaussian prior over the robot poses, $\cProb{\cposes}$, with precision $\epsilon$, yielding the information matrix $\epsilon\cMat{I}$, where $\cMat{I}$ is the identity matrix. Equivalent to (\ref{eq:ent-sum-probs}), the marginal posterior information matrix over the poses $\cInfo{\cposes}^S$ can be written in terms of a prior information matrix plus a term contributed by each selected map point $\cInfo{\cposes}^i$, or 
         \begin{equation}\label{eq:ent-sum}
            \cInfo{\cposes}^S = \epsilon\cMat{I}+\sum_{i \in S} \cInfo{\cposes}^i.
        \end{equation}
        
        The information matrix from each selected map point $i$, or $\cInfo{\cposes}^i$, is given by the marginal of the joint map point information matrix, or $\cInfo{\cposes,\cmappoint{i}}$. The joint map point information matrix is given by the sum of all the measurements of the map point $\cmappoint{i}$, or 
        \begin{equation}\label{eq:jointlandmark}
            \cInfo{\cposes,\cmappoint{i}} = \sum\limits_{j=1}^{t} \cTranspose{\cMat{J}_{ij}}\cMat{\Omega}_{ij}\cMat{J}_{ij},
        \end{equation}
        where $\cMat{J}_{ij}$ and $\cMat{\Omega}$ are the SLAM measurement Jacobian and measurement noise defined in subsection \ref{subsec:vslam}. Before we obtain the marginal information matrix, we first partition the joint information matrix $\cInfo{\cposes,\cmappoint{i}}$ into sub-matrices, or 
            \begin{equation}\label{eq:schur-blocks}
                    \cInfo{\cposes,\cmappoint{i}}  = 
                    \begin{pmatrix}
                        \cMat{C}_i & \cMat{B}_i \\ 
                        \cMat{B}_i^{\text{T}} & \cMat{P}_i 
                    \end{pmatrix} 
            \end{equation}
        where $\cMat{C}_{i}$ is the block associated with all the robot poses, $\cMat{P}_i$ is the block associated with map point $i$ and $\cMat{B}_i$ as the off\hyp{}diagonal entries. With the sub-matrix definitions in (\ref{eq:schur-blocks}), $\cInfo{\cposes}^i$ is given by using the Schur complement on $\cInfo{\cposes\cmappoint{i}}$, or 
        \begin{equation}\label{eq:schur}
            \cInfo{\cposes}^i = \cMat{C}_i - \cMat{B}_i^{}\cMat{P}_i^{-1}\cMat{B}_i^{\text{T}}.
        \end{equation}
        This concludes the baseline SLAM utility definition as (\ref{eq:schur}) can be substituted into (\ref{eq:ent-sum}) and (\ref{eq:slam-def}) for each map point. 
        
        We consider the computational complexity of evaluating the baseline SLAM utility. For greedy algorithms, it is necessary to calculate the marginal gain of a map point at each iteration. The marginal gain of map point with index $i$ for the baseline SLAM utility is given by 
        \begin{equation}\label{eq:slam-marg-gain}
            f_{\text{SLAM}}(m_i|S) = \frac{1}{2} \log(|\cInfo{\cposes}^S + \cInfo{\cposes}^i |) - \frac{1}{2} \log(|\cInfo{\cposes}^S|).
        \end{equation}

    To evaluate the marginal gain in (\ref{eq:slam-marg-gain}), we need to compute the determinant of a $6t$ by $6t$ matrix, where $t$ represents the number of robot poses in the trajectory. Using Cholesky decomposition for this calculation takes $O(t^3)$ if the decomposition is calculated from scratch. If we instead store the Cholesky decomposition of the currently chosen map point set, or $\cInfo{\cposes}^{S}$, we can use the Cholesky update routine to calculate the determinant of $\cInfo{\cposes}^{S \cup \{m_i \}}$ which takes $O(t^2)$, assuming $\cInfo{\cposes}^{i}$ is low rank. This approach is similar to updating the Cholesky decomposition for incremental SLAM approaches between timesteps (see ISAM 2 \cite{kaess2012isam2}, SLAM++ \cite{ila2017slam++}) or similar to selecting poses for SLAM in \citet{zhao2020good}. Our software implementation uses the CHOLMOD \cite{chen2008algorithm} sparse Cholesky update implementation.

Given the complexity of the marginal gain, the complexity of maximising the SLAM utility with the stochastic greedy algorithm using Cholesky updates takes $O(t^2n)$ time. Since a bounded number of new map points are detected at each timestep, the total number of map points $n$ is at most proportional to the trajectory length $t$. Therefore, the resulting asymptotic worst-case complexity of this approach is $O(t^3)$. This is noteworthy since it is equivalent to the worst-case complexity of SLAM estimation using bundle adjustment, which implies similar asymptotic scaling. However, for practical execution times, it is important to note that there is typically an order of magnitude more map points than poses in the trajectory. Therefore, we expect this approach to be very computationally expensive in practice.

Calculating the determinant requires, at worst, $O(t^2)$ memory. Storing the map point observations and the associated features takes $O(n|X_i|)$ memory, where $|X_i|$ is the upper bound of the number of frames from which a map point is visible. Caching the intermediate information matrices from (\ref{eq:schur}) requires $O(n|X_i|^2)$ memory. As a result, we expect this approach to use a large amount of memory in addition to being computationally expensive.

    \subsection{Utility based on a Localisation Approximation}\label{subsec:local}
         The previous subsection proposed a utility based on the SLAM posterior, but this baseline utility was expensive to evaluate. This subsection proposes a novel utility that improves the scalability of the baseline SLAM approach by approximating the SLAM posterior with a set of independent localisation problems. This approximation results from the assumption that perfect knowledge of the map points has a negligible effect on the SLAM posterior, or  
        \begin{equation}\label{eq:local-assumption}
                    \cProb{\cposes|\cobservations^{S}}  
        \approx \cProb{\cposes|\cobservations^{S},\cmappoint{S}}.
        \end{equation}
        This approximation is reasonable if there is very little uncertainty in the map point positions. We expect this simplifying assumption will overvalue the utility of map points with a limited number of observations or poor viewpoint coverage. The impact of this approximation when using this utility with greedy algorithms will depend on the utility of the resulting greedy selection of map points. If using the approximate utility function still accurately ranks the relative utility of map points, the greedy algorithm will continue to select sets with high utility despite using an approximate utility function.  
    
    We use the previous assumption from the SLAM utility that the prior distribution over the robot trajectory is independent and identically distributed allow us to factorise the posterior distribution into factors associated with each pose as  
        \begin{equation}\label{eq:local-independant}
            \cProb{\cposes|\cobservations^{S},\cmappoint{S}}  = \eta\prod\limits_{j=1}^{t}  \cProb{\cpose{j}} \cProb{\cobservation{j}^{S} |\cpose{j},\cmappoint{S}}, 
        \end{equation}
        where $\eta$ is a normalising constant.

Each factor in the product in (\ref{eq:local-independant}) only involves one pose and is independent of other poses due to the assumptions made previously. This simplifies the calculation of the SLAM utility in (\ref{eq:slam-def}), to the sum of terms originating from independent localisation problems, or 
\begin{equation}\label{eq:local}
    f_{\text{local}}(S) = \sum\limits_{j=1}^{t} \log(| \cInfo{\cpose{j}|\cmappoint{}}^S|) - c_{\emptyset},
\end{equation}
where $\cInfo{\cpose{j}|\cmappoint{}}^S$ is the information matrix for $\cpose{j}$ due to the localisation approximation. The information matrix $\cInfo{\cpose{j}|\cmappoint{}}^S$ can be calculated in terms of the selected map point measurements with measurement Jacobian $\cMat{A}_{ij}$ and measurement noise $\cMat{\Omega}_{ij}$, or
\begin{equation}\label{eq:local-mats}
     \cInfo{\cpose{j}|\cmappoint{}}^S =\epsilon \cMat{I} + \sum_{i\in S} \cMat{A}_{ij}^{}\cMat{\Omega}\cMat{A}_{ij}^\text{T}.
\end{equation}

We now consider the complexity of evaluating the marginal gain for the localisation approximation. We define $X_i$ as the set of poses from which map point $\cmappoint{i}$ is visible. By substituting the localisation utility in (\ref{eq:local}) into the marginal gain definition in (\ref{eq:marggain}),
the marginal gain simplifies to only the terms related to poses from which $\cmappoint{i}$ was visible, or 
\begin{equation}\label{eq:local-margain}
    f_{\text{local}}(m_i|S) = \sum\limits_{j \in X_i } \log(| \cInfo{\cpose{j}|\cmappoint{}}^{S \cup \{m_i\}}|)-\log(| \cInfo{\cpose{j}|\cmappoint{}}^S|).
\end{equation}
Note that the determinant of each localisation sub\hyp{}problem is of constant size. While evaluating the localisation utility in (\ref{eq:local}) takes $O(t)$ time, evaluating the marginal gain in (\ref{eq:local-margain}) takes only $O(|X_i|)$ time. This reduces to $O(1)$ if we assume the number of poses from which map points are visible, or $|X_i|$, is bounded. This is significantly less expensive than the  $O(t^2)$ complexity of evaluating the marginal gain for the full SLAM utility. Maximising the localisation utility function with the lazy greedy algorithm, therefore, has a worst\hyp{}case computational complexity of $O(n^2)$.

To implement this approach, the current information matrix of each frame, or $\cInfo{\cpose{j}|m}^S$ must be stored and updated whenever a map point is added. The current information matrix for each frame is initialised which the prior matrix, or 
    \begin{equation}
        \cInfo{\cpose{j}}^{\emptyset} = \epsilon \cMat{I}. 
    \end{equation}
    Whenever a map point is added to the currently selected set, the stored matrices for each frame is also updated, or 
    \begin{equation}
        \cInfo{\cpose{j}}^{S \cup \{m_i\}}  = \cInfo{\cpose{j}}^{S} +  \cMat{A}_{ij}^{}\cMat{\Omega}\cMat{A}_{ij}^\text{T}.
    \end{equation}
    
    Storing the $6 \times 6$ information matrix for each frame takes $O(t)$ memory. It is possible to speed up the evaluation of the marginal gain in (\ref{eq:local-margain}) by storing the intermediate information matrices  $\cMat{A}_{ij}^{}\cMat{\Omega}\cMat{A}_{ij}^\text{T}$ associated with map point observation. Storing these intermediate matrices requires $O(n|X_i|)$ memory and is, therefore, not recommended for memory-constrained applications. 
 
    \subsection{Utility based on a Stereo Odometry Approximation}\label{subsec:odom}
       Subsection \ref{subsec:local} proposed a utility based on approximating the SLAM posterior with a set of localisation problems. This subsection instead proposes a novel approximation that approximates the posterior with a set of independent stereo odometry problems. 

       The entropy of the posterior distribution over poses can be expanded using the chain rule to the sum of entropy terms conditioned on all observations and previous robot poses in the chain, or 
         \begin{equation}\label{eq:entropy-chain}
            \cEntropy{\cposes|\cobservations^{S}} = \cEntropy{\cpose{1}|\cobservations^{S}} + \cEntropy{\cpose{2} |\cpose{1},\cobservations^{S}} + \dots +
            \cEntropy{\cpose{t}|\cpose{1:t-1},\cobservations^{S}}.
         \end{equation}
         The odometry utility uses the general property of entropy that conditioning on fewer variables always increases the entropy. For the odometry utility, each term within (\ref{eq:entropy-chain}) is approximated by conditioning only on the variables involved in a pair of timesteps. Note that we cannot use consecutive poses since keyframes are only inserted when exploring novel regions, and consecutive poses might not share map points. Instead of using consecutive poses, this paper uses the simple heuristic of pairing the pose $\cpose{j}$ with the frame with the highest number of co-visible map points $\cpose{p_j}$ with a lower index, or $p_j < j$. Despite the simplicity of this approach, we found it to work well in practice. The entropy of each pose term in the chain is approximated by conditioning on the pair of poses and the associated observations from those poses, 
         \begin{equation}\label{eq:entboundodom}
            \cEntropy{\cpose{j}|\cpose{1:j-1},\cobservation{1:t-1}^S} \approx \cEntropy{\cpose{j}|\cpose{p_j,j},\cobservation{p_j,j}^{S}}.
        \end{equation} 
        
        Substituting (\ref{eq:entboundodom}) into the sum in (\ref{eq:entropy-chain}) yields a bound on the posterior entropy using the sum of pose pairs, or
        \begin{equation}\label{eq:odom-ent}
            \cEntropy{\cposes|\cobservations^{S}} \approx \cEntropy{\cpose{1}} + \sum\limits_{j=2}^{t} \cEntropy{\cpose{j}|\cpose{p_j},\cobservation{p_j,j}^S}.
        \end{equation}
        The information gain using the SLAM posterior is the difference between in $\cEntropy{\cposes}$ and $\cEntropy{\cposes|\cobservations^S}$. Substituting (\ref{eq:odom-ent}) into this equation yields the map point utility based on visual odometry, or 
        \begin{equation}\label{eq:odom-eq}
            f_{\text{odometry}}(S) 
          = c_{\emptyset} + \sum\limits_{j=2}^{t} \log( |\cInfo{\cpose{j}|\cpose{p_j} }^S |),
        \end{equation}
    where $\cInfo{\cpose{j}|\cpose{p_j} }^S$ is the information matrix of the conditional distribution over $\cpose{j}$, conditioned on the paired pose $\cpose{p_j}$ and the measurements between those poses. This odometry utility formulation is a lower bound to the information gain from map point measurements using the SLAM posterior. It is also worth noting that this approach implicitly only evaluates stereo odometry information by only considering a pair of poses.

     As with the localisation approximation, this introduces an additional simplifying assumption to the SLAM utility. By only considering the stereo odometry problems, this utility will undervalue map points observed from multiple frames and does not consider map point observations where the disparity was too low and which was therefore modelled as monocular observations. The impact of this approximation will depend on how it affects the selection of the greedy algorithm.

    As noted in Subsection \ref{sec:framework}, the greedy algorithms requires calculating the marginal gain at each iteration. The marginal gain of odometry utility for a map point $m_i$ that only is visible from a limited set of poses $X_i$, simplifies to 
    \begin{equation}
         f_{\text{odometry}}(m_i|S) =  \sum\limits_{j \in X_i} \log( |\cInfo{\cpose{j}|\cpose{p_j} }^{S \cup \{ m_i \}} |) - \log( |\cInfo{\cpose{j}|\cpose{p_j} }^S |).
    \end{equation}\label{eq:odom-eq-marg}
    This yields a similar form as the marginal gain for localisation utility in (\ref{eq:local-margain}). If the size of $X_i$ is bounded, evaluating the marginal gain for the odometry utility has a computational complexity of $O(1)$. Maximising this utility with lazy greedy and stochastic greedy therefore has a complexity of $O(n^2)$ and $O(n)$, respectively. 
    
    The odometry utility requires storing the information matrices associated with each frame, which requires $O(t)$ memory. From both a computation and memory usage perspective, the odometry utility is equivalent to the localisation utility and serves as another approximate alternative to the SLAM utility from Subsection \ref{subsec:entropy}. 

\section{Combined Utility Formulations for Improved Loop-closure Detection Performance}\label{sec:loop}
   
The previous section proposed information\hyp{}theoretic models for map point utility based on the estimation performed in the SLAM back-end. We expect that a definition of map point utility that considers both the SLAM front-end and back-end should outperform approaches that only consider the back-end estimation problem. Consequently, this subsection proposes augmenting an information-theoretic utility with an additional term to improve its ability to capture aspects of the front-end, specifically loop closure detection. This is used for a combined utility function can be formulated that is the weighted sum of the information\hyp{}theoretic utility function and the loop\hyp{}closure utility function.

This front-end task of loop-closure detection is an essential part of a SLAM implementation affected by map point selection not modelled by our previous utility functions. Loop detection is typically implemented as a two-stage process: first, loop-closure candidates are identified using appearance-based techniques (e.g. \cite{cummins2008fab}). Secondly, these loop-closure candidates are verified by performing feature matching. Candidates with sufficient feature matches are accepted, while others are rejected as outliers. Map point selection will result in fewer map points available for feature matching during loop-closure detection, which can negatively impact performance. 

In existing work, the problem of selecting map points for localisation in maps generated offline is often formulated as variations of the set multi-cover problem; that is, selecting the minimum number of map points to ensure that each camera still contains a desired threshold of selected map points. One expects that due to the success of these map point selection approaches for localisation, they will continue to facilitate good performance for the same localisation algorithms used in visual SLAM implementations.

We make the simplifying assumption that we can obtain the subset of keyframes or poses where it is necessary to be able to perform loop\hyp{}closures. In scenarios where a planner is available, this information could be extracted by evaluating the future trajectory. Alternatively, such information could be obtained from application-specific information. For some applications, it would be feasible to extract the likelihood or probability that a loop\hyp{}closure occurs in a specific scene (for example, by training a neural network-based classifier). In cases where such information is not available, this version of the algorithm serves as a comparison point against which to evaluate other algorithms.

For our loop-closure utility to be compatible with the greedy formulation in Section \ref{sec:framework}, function must be monotone submodular. We use the set coverage of the loop-closure frames $L$ as a utility, which is monotone and submodular. For each frame $j$, we define the map points visible in that frame as $V_j$. The loop closure coverage is then given by:
\begin{equation}\label{eq:cover}
    f_{\text{cover}}(S) = \frac{1}{|L|}\sum\limits_{j \in L} \min \left(S \cap V_j, b_{\text{cover}} \right),
\end{equation}
where $b_{\text{cover}}$ is the chosen coverage parameter and should be adjusted based on the number of map points needed to reliably perform loop-closures.

The complexity of evaluating this utility is $O(L)$. Additionally, it is easy to show that the marginal gain is proportional to the number of co-visible loop-closure frames.

We choose to combine the odometry utility from Subsection \ref{subsec:odom} with the utility in (\ref{eq:cover}) as a weighted sum, but this could similarly have been applied to the localisation approach. This choice is based on preliminary experimentation, which showed that the odometry approach often obtained good trajectory accuracy. For the relative weighting of odometry and loop utility, it is important to consider that the utility functions are on scalled different. We found that using equal weighting after normalising the respective utilities with the utility obtained for selecting all map points, or $V$, worked well despite the heuristic nature of this approach. The normalised sum of our utility functions $f_{\text{odom}}(S)$ and $f_{\text{cover}}(S)$ yields the combined utility function: 
\begin{equation}\label{eq:odom+cover}
                f_{\text{odom+cover}}(S) = \left(\frac{1}{f_{\text{odom}}(V)}\right)f_{\text{odom}}(S)  + \left(\frac{1}{f_{\text{cover}}(V)}\right)f_{\text{cover}}(S).
\end{equation} 

An alternative approach we experimented with was using probabilistic models for coverage of the loop-closure frames. However, we found that probabilistic approaches did not necessarily outperform the utility presented here. When using the combined utility such as in (\ref{eq:odom+cover}), the information-theoretic term already biases map point selection towards stable map points which was informative in the past.

This section proposed the inclusion of additional information regarding loop closures to the odometry utility. We expect that this combined approach will improve performance for maps where using the odometry utility resulted in selecting too few map points in the provided loop-closure locations.

\section{Experimental Setup}\label{sec:setup}
This section presents the experimental setup used to compare the impact of different map point selection approaches on the accuracy of SLAM estimation. We choose ORB-SLAM 2 as a representative sparse visual SLAM algorithm but introduce modifications and a test procedure that selects map points offline.

    Using a SLAM software implementation allows testing on practical data and tests how map point selection impacts the overall SLAM implementation. ORB-SLAM 2 has state-of-the-art accuracy, is validated performance on a range of datasets, and is open-source \cite{mur2017orb,ORBSLAM3}. We found that ORB-SLAM 3 performs worse than the original ORB-SLAM 2 algorithm in the KITTI dataset when using only stereo cameras (as shown in Subsection \ref{subsec:base-kitti}). ORB-SLAM 3 includes improved support for different sensors, such as inertial measurement units, which we do not use here. For these reasons, this paper uses ORB-SLAM 2 despite the availability of a new version. For the scope of this work, we choose to use the stereo vision datasets with ground truth information already compatible with the ORB-SLAM 2 algorithm: The KITTI odometry and EuRoC MAV datasets. ORB-SLAM 2 also supports the TUM-RGBD dataset by converting RGBD measurements into pseudo-stereo measurements, but we did not use this dataset or functionality, as would require adapting the proposed information-theoretic approaches for RGBD-sensors.

ORB-SLAM 2 has to important challenges when used for evaluating map point selection approaches: the first is that the algorithm is stochastic and can exhibit significant variation between runs of the algorithm. This variation can prevent results in different map points between runs, which makes it challenging to compare the relative performance of map point selection approaches on the same data. The second challenge is that some approaches (such as the SLAM utility function from Subsection \ref{subsec:entropy}) are unsuitable for online use.

The rest of the Section is organised as follows: we discuss SLAM evaluation metrics in Subsection \ref{sec:rpe-metrics}. Subsection \ref{subsec:orb-mod} highlights the relevant properties and presents modifications to the algorithm for evaluating map point selection approaches. We present a offline test procedure in Subsection \ref{subsec:testproc}; this test procedure requires separating trajectories from the datasets into ``input'' and ``test'' trajectories. Baseline map point selection approaches are detailed in Subsection \ref{subsec:baselines}. We detail how we chose the previously mentioned trajectories for the KITTI and EuRoC datasets in Subsections \ref{subsec:base-kitti} and \ref{subsec:base-euroc}, respectively.

\subsection{SLAM Evaluation Metrics}\label{sec:rpe-metrics}
   This paper uses the accuracy of a SLAM algorithm to evaluate different map point selection approaches. The accuracy of SLAM algorithms is typically evaluated using the estimated trajectory and comparing it against the ground truth trajectory. The estimated SLAM trajectory for SLAM is typically evaluated using either the absolute pose error (APE) or relative pose error (RPE) \cite{sturm2012benchmark}. APE evaluates errors by comparing first aligning the estimated trajectory with the ground truth trajectory, and then evaluating the error between estimated poses. RPE instead compares the relative poses a specified distance along the trajectory. For the KITTI dataset, we evaluate RPE over distances of 100 to 800 metres \cite{Geiger2012CVPR}. 
   We found that RPE is not useful for evaluating map point selection approaches on the smaller scale EuRoC dataset and therefore, used APE for EuRoC. This decision is motivated Subsection \ref{subsec:base-euroc}, showing the RPE results using the RPE definition of \citet{sturm2012benchmark}.

     In addition to the trajectory error metrics, we evaluate the performance of the loop\hyp{}closure module. Evaluating the performance of this module assists in interpreting the trajectory accuracy results of map point selection, as identifying loop closures have a vital role in trajectory estimation. Identifying a loop closure is often modelled as a classification problem; that is, classifying a frame as a loop closure with a previous frame or a novel location. Recall for classifying loop\hyp{}closures is the fraction of true loop closures correctly labelled or 
     \begin{equation}
                    \text{Recall} = \frac{\text{Correctly labelled loop closures}}{\text{Ground truth loop closures}}.
            \end{equation}
     This is in contrast to precision, which would be the ratio of correctly labelled frames to the frames that are labelled as loop closures. It is common to evaluate both precision and recall for classification problems, as these are competing objectives. For SLAM implementations, loop\hyp{}closure algorithms are designed to have very high precision at the expense of recall. While not shown here, it was found that recall performance can drop significantly for the map point selection approaches, and precision is mostly unaffected. This paper consequently only reports recall for loop\hyp{}closure performance for map point selection, as precision remained at 100\%.

\subsection{Modifications to ORB-SLAM 2}\label{subsec:orb-mod}
While ORB-SLAM 2 is useful for evaluating map point selection approaches, some properties stemming from its software implementation present challenges when evaluating map point selection. This subsection proposes modifications to the ORB-SLAM 2 algorithm to facilitate the evaluation of map point selection approaches.

Four properties of the original ORB-SLAM 2 are not suited for evaluating map point selection approaches. Firstly, ORB-SLAM 2 exhibits significant variation due to stochastic algorithms and multi-threaded implementation that prevents testing selection approach on the same map. Secondly, the algorithm relies on pose-graph optimisation. This optimisation does not include map points and diminishes our ability to evaluate map point selection approach by the impact on SLAM estimation. Thirdly, the online multi-threaded nature prevents the testing of offline map point selection approaches. Lastly, current heuristics for removing redundant keyframes could be better adjusted for the sparse maps resulting from map point selection. 

 We modified the ORB-SLAM 2 algorithm to save and load maps offline and tested map point selection approaches on the maps stored offline. This allows testing map point selection approaches on the same map, reducing the impact of variation on the algorithm and allowing a more fair comparison. This modification also allows testing more computationally expensive map point selection approaches unsuitable for online use, such as the SLAM utility from Subsection \ref{subsec:entropy}.
       
 Another change was to increase the number of bundle adjustment iterations and modify the algorithm to wait for bundle adjustment to finish before processing new measurements. This change allows estimation to converge more reliably, which reduces variation in estimated trajectory accuracy. This change places a greater emphasis on selecting a set of map points that allows for accurate SLAM estimation and less of an emphasis on the accuracy of the pose-graph heuristic. A side effect of this is that it reduces the real-time nature of the algorithm. This change is aimed at improving the ability to differentiate map point selection approaches.   
 
 The ORB-SLAM 2 algorithm is also changed to avoid problems with the incorrect removal of keyframes. This heuristic is altered not to remove keyframes affected by map point selection. Introducing map point selection into other visual SLAM algorithms would require a similar revision of heuristics that are no longer appropriate in the presence of map point selection.

 Our modified version of ORB-SLAM 2, as described by this subsection, is publically available\footnote{https://github.com/ChristiaanM/ORB\_SLAM2\_selection}.
   
\subsection{Test Procedure for Evaluating Map Point Selection Approach}\label{subsec:testproc}
 Subsection \ref{subsec:orb-mod} presented the changes made to ORB-SLAM 2 to allow offline evaluation of map selection approaches, in order for different approaches to use the same map and to evaluate approaches not suitable for online use.

\begin{figure}[htb]
            \centering
            \def\svgwidth{\linewidth}
            \fontsize{9pt}{10pt}
            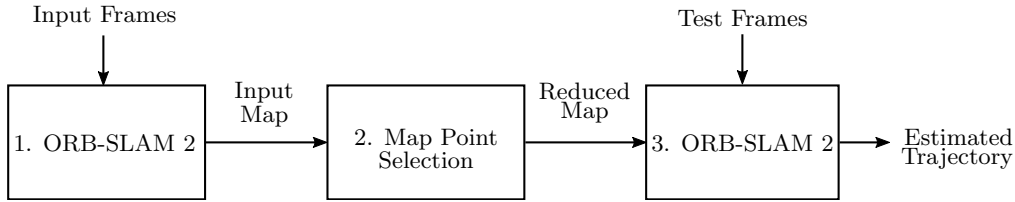
            \caption{
                This flow chart shows the sequential steps used to evaluate map point approaches with offline selection. An initial input portion of a trajectory is used with a modified ORB-SLAM 2 implementation to generate an input map. A map point selection approach is then applied offline to the input map to simplify the map. Lastly, the ORB-SLAM 2 algorithm is started again using the remaining test portion of the trajectory and the reduced map as input.    
            }
            \label{fig:testprocedure}
        \end{figure}

 A flow chart of the test procedure is shown in Figure \ref{fig:testprocedure}. The test procedure segments trajectories from SLAM datasets into two consecutive parts: the input trajectory and the test trajectory. The input trajectory is used with the modified version of ORB-SLAM 2 to create an input map, which is stored. Different map point selection approaches are then applied on the same input map. The reduced maps generated by the different selection approaches then replace the input map in the modified ORB-SLAM 2, after which SLAM is performed on the remaining test trajectory. We detail how we select these trajectories from the KITTI and EuRoC datasets in Subsections \ref{subsec:base-kitti} and \ref{subsec:base-euroc}, respectively.
 
 This procedure avoids penalising expensive selection approaches based on the effect on the timing of threads. It also allows the testing of selection approaches on the same input maps, despite the stochastic nature of the ORB-SLAM 2 algorithm.

Each map point selection approach is evaluated on all the chosen trajectories of a dataset. For each trajectory, the process of generating an input map from the input trajectory is repeated five times; this produces five input maps for each trajectory to compensate for the stochastic nature of the ORB-SLAM 2 algorithm. By using the same set of input maps for all approaches, we reduce potential bias due to differences in the maps between approaches. Also, the ability to capture the variation in performance is improved by testing with a group of input maps. The impact of the variation in maps resulting from the stochastic nature of the algorithm is shown in greater detail in \ref{sec:kitti-map-variation}.

A selection approach is applied to each of the five input maps. Each reduced map resulting from the selection is then evaluated twice on the remaining trajectory. This results in 10 samples for each trajectory per approach at a tested map point budget. For stochastic selection approaches (such as random selection described in Subsection \ref{subsec:baselines}), we apply each selection approach five times instead of only once per map. This results in 50 samples for each trajectory at a tested map point budget.

\subsection{Baseline Map Point Selection Approaches}\label{subsec:baselines}
    
    In this subsection, we review some baseline map point selection approaches. These baseline approaches are used as a important comparison point in Section \ref{sec:experiment} and can provide insight into the measurable impact of map point selection for a chosen trajectory.
    
    As an alternative to the different map point selection approaches proposed in this paper, there are several important baselines. The first baseline is including all the map points in the map (or not using map point selection). We label this approach ``full map''. Generally, we expect to achieve the best trajectory using all the available map points. The aim of the proposed approaches is to offer comparable accuracy at lower map sizes. 
    
    Another alternative is selecting no map points (i.e. discarding the input map). Since the ORB-SLAM 2 algorithm resets if it fails to localise itself in an existing map, evaluating the trajectory for no map points is not meaningful. Instead, we include all the map points visible in the last keyframe and label this approach ``empty map''. This is equivalent to starting a new map with the test trajectory  and consequently prevents any loop-closures with the input trajectory. 
    
    The last important baseline to compare a map selection approach with a given map point budget is randomly selecting a set of map points. We label this approach ``random''. For a given map point selection approach to be useful, we expect it to compare favourably to random selection using the same number of map points. 
   
\subsection{The KITTI Dataset Trajectories}\label{subsec:base-kitti}
    The previous subsections covered both the experimental setup and metrics used to evaluate map point selection algorithms. This section discusses the KITTI dataset for evaluating map point selection algorithms using the proposed testing setup. Not all trajectories in this dataset are useful for evaluating map point selection algorithms. This subsection chooses a subset of the KITTI trajectories and evaluates the baseline performance when using these chosen trajectories with the modified ORB-SLAM 2 version from Subsection \ref{subsec:orb-mod}.
    
   The KITTI dataset features twenty trajectories of a vehicle driving in an urban environment. These trajectories range up to several kilometres per trajectory. Only the first ten trajectories include publicly available ground truth information; therefore, only these trajectories are considered for evaluating map point selection. For trajectories that do not revisit regions of the map, the selection or retention of map points has a minimal effect on trajectory accuracy. In contrast, it is essential to maintain map points for accurate estimation during large loop closures. This paper consequently chooses the subset of trajectories from the KITTI dataset that includes loop closures reliably detectable by the ORB-SLAM 2 algorithm (without selection). The algorithm cannot detect loop closures from significant variations in viewpoint, such as for the 08 trajectory. The ORB-SLAM 2 algorithm also fails to reliably detect the loop closures in the 09 trajectory due to limited overlap \cite{mur2017orb}. After this elimination, the following KITTI trajectories are chosen for evaluation: 00, 02, 05, 06, and 07. 

    \begin{figure}[tbp]
            \centering
            \input{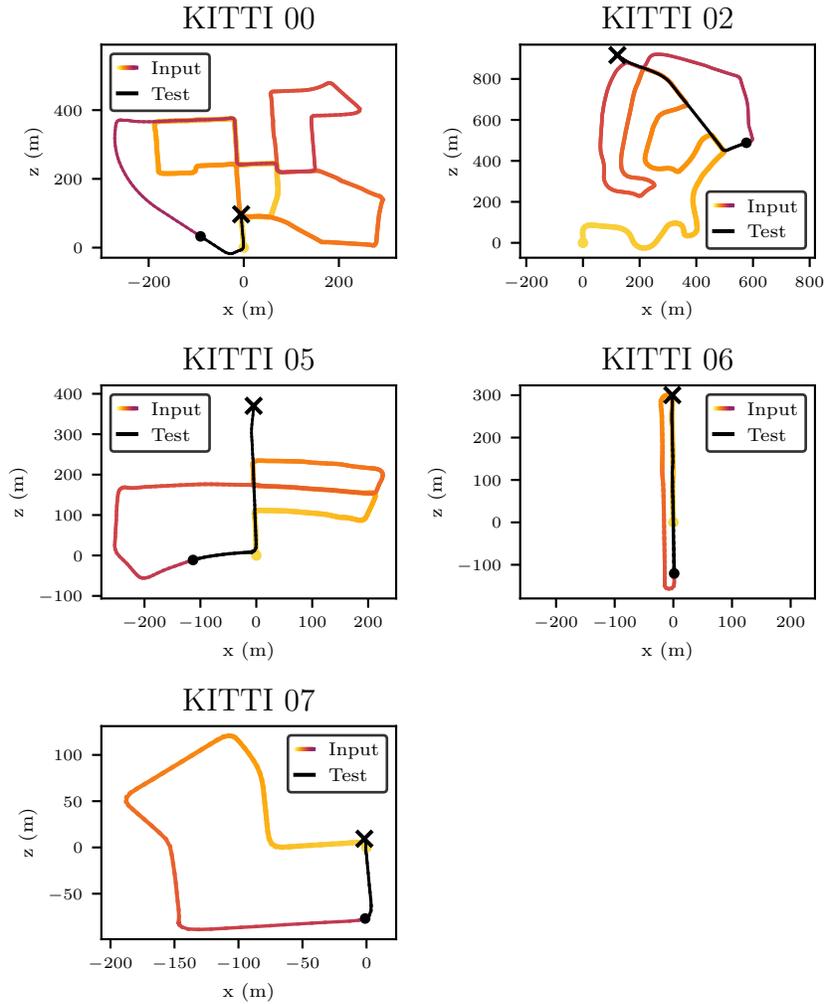}
            \caption{
            The chosen trajectories for the KITTI dataset for evaluating map point selection approaches. The KITTI trajectories labelled 00, 02, 05, 06 and 07 are divided into a test and input trajectory, where the map from the initial input trajectory (progress along this trajectory is visualised with a colour gradient) is reduced using a map point selection approach and tested on the remaining test trajectory (from black dot to the cross). The origin of the trajectories is the camera location of the starting position of the ground truth trajectory (yellow dot). }
            \label{fig:input-traj}
    \end{figure}

   Each of these trajectories is separated into a input and test trajectory for the test procedure described in Subsection \ref{subsec:testproc}. Since loop closures play an essential role in SLAM estimation, we choose to separate the trajectories before large loop closures. Our choices for the number of frames of the input trajectories are as follows: 4300 frames for KITTI 00, 4100 for KITTI 02, 2200 frames for KITTI 05, 750 frames for KITTI 06 and 900 frames for KITTI 07. The path of trajectories are also shown in Figure \ref{fig:input-traj}.

      We evaluate the proposed setup's relative pose error using the ``full map'' baseline from Subsection \ref{subsec:baselines} to characterise the modified ORB-SLAM's performance using this test procedure. For comparison, we include the original ORB-SLAM 2 and ORB-SLAM 3 algorithms. These respective algorithms are evaluated on each of the chosen KITTI trajectories. Table \ref{fig:base-acc} shows the mean and standard deviation of the relative pose errors. It should be noted that the accuracy of the ORB-SLAM algorithms is hardware-dependent.
 
        \begin{table}[tbp]
                    \centering
                    \begin{tabular}{lccc}
                        \toprule
                    Trajectory	&  ORB-SLAM 3 & ORB-SLAM 2 & Modified ORB-2\\
                    \midrule
                    00 & 0.717 $\pm$ 0.011 & 0.701 $\pm$ 0.008 & 0.666 $\pm$ 0.005 \\
                    02 & 0.774 $\pm$ 0.016 & 0.776 $\pm$ 0.013 & 0.723 $\pm$ 0.010 \\
                    05 & 0.430 $\pm$ 0.020 & 0.404 $\pm$ 0.015 & 0.408 $\pm$ 0.005 \\
                    06 & 0.538 $\pm$ 0.035 & 0.513 $\pm$ 0.113 & 0.476 $\pm$ 0.012 \\
                    07 & 0.471 $\pm$ 0.036 & 0.512 $\pm$ 0.044 & 0.465 $\pm$ 0.033 \\
                        \bottomrule
                    \end{tabular}
                    \caption{The variation of RPE (\%) for different versions of ORB-SLAM on the chosen stereo trajectories of KITTI. Modifications to the original ORB-SLAM 2 algorithm reduced variance at the expense of additional delays. The mean and standard deviation of the RPE are reported for the original ORB-SLAM 2 and ORB-SLAM 3 algorithm, as well as the modified ORB-SLAM 2 version used in this paper to evaluate map point selection. The quoted results are from 20 samples for each corresponding algorithm. 
                     }
                    \label{fig:base-acc}
        \end{table}

        As shown in the results from Table \ref{fig:base-acc}, the modified ORB-SLAM 2 algorithm discussed in Subsection \ref{subsec:orb-mod} resulted in a reduced standard deviation in trajectory error when compared to other implementations of ORB-SLAM. A side effect of these changes is the improved trajectory accuracy, at the expense of additional delays in the algorithm. These changes were motivated to improve the ability to more consistently evaluate the impact of map point selection of trajectory accuracy. Despite the improved trajectory accuracy, the additional delays might not be appropriate for all settings.
     
        The results further show that ORB-SLAM 3 performs worse than the ORB-SLAM 2 algorithm on the KITTI dataset. ORB-SLAM 3 algorithm modified several heuristics related to loop closures and no longer performs bundle-adjustment using the full map for larger maps and instead only relies on the pose graph estimation. We suspect that these changes could have resulted in reduced trajectory accuracy for the KITTI dataset. 
        
        For a map point selection approach to do well for the KITTI trajectories, the approach will have to efficiently reduce maps without negatively impacting either the SLAM estimation or the front-end algorithms. Notably, the selection approach will have to select sufficient map points to continue identifying loop\hyp{}closures and accurately performing SLAM estimation.

    \subsection{The EuRoC Dataset Trajectories}\label{subsec:base-euroc}
        The EuRoC dataset \cite{burri2016euroc} is also chosen to evaluate map point selection approaches. These trajectories include challenging dynamic motion of a micro air vehicle in a smaller environment than the KITTI dataset. These trajectories feature one of two environments: either the machine hall or the Vicon room. The different trajectories in these environments feature varying degrees of challenging motion for trajectory estimation. Accurate ground truth is provided using a laser tracker for the machine hall trajectories, while the Vicon room uses a Vicon motion capture system.
        
        Unlike the KITTI dataset, the trajectories in this dataset often repeatedly cover the same small-scale environment and rely less on a long-term appearance\hyp{}based loop\hyp{}closure heuristic. Therefore, we cannot use large-scale loop closures to guide the choice of how to split the trajectory into an input and test trajectory or as a means to eliminate trajectories that are not useful for evaluating map point selection. We instead choose to use the simple scheme of separating the trajectory into two equal halves for the test and input trajectory.
        
       Using the aforementioned choice of separating the trajectories, we compare the performance between the ``empty map'' and ``full map'' baselines described in \ref{subsec:baselines}. If selecting no map points can provide a comparable trajectory accuracy to the full map, we do not expect the chosen combination of trajectories and metric to be useful for evaluating map point selection approaches. The absolute and relative trajectory errors of these baseline approaches are reported for the respective trajectories in Table \ref{fig:base-acc-euroc}. 
       
        \begin{table}[htbp]
                    \centering
                    \begin{tabular}{lr|rr|rr}
                        \toprule
                                  & & \multicolumn{2}{c|}{APE (m)}  & \multicolumn{2}{c}{RPE (m)}    \\                                 
                        Trajectory   & Frames & Empty map               & Full map                & Empty map               & Full map    \\            
                        \midrule
MH01 & 1841           & 0.05                & 0.04                & 0.56                & 0.56                \\
MH02 & 1520        & 0.04                & 0.02                & 0.63                & 0.63                \\
MH03 & 1350         & 0.03                & 0.03                & 1.07                & 1.07                \\
MH04 & 1017         & 0.15                & 0.13                & 1.00                & 1.00                \\
MH05 & 1137         & 0.13                & 0.06                & 0.94                & 0.94                \\
V101 & 1456         & 0.04                & 0.03                & 0.33                & 0.33                \\
V102 & 885          & 0.07                & 0.02                & 0.78                & 0.77                \\
V103 & 1075         & 0.17                & 0.04                & 0.62                & 0.60                \\
V201 & 1140         & 0.07                & 0.04                & 0.36                & 0.36                \\
V202 & 1174         & 0.05                & 0.03                & 0.71                & 0.71                \\

                        \bottomrule
                    \end{tabular}
                    \caption{
Evaluation of the impact of map point selection on the EuRoC trajectories. Each trajectory in the EuRoC dataset is split into equal input and test portions. 
This evaluation is repeated 10 times and the median trajectory errors are reported. The trajectory accuracy of keeping all the map points is compared to discarding all the map points except those in the last keyframe of the input trajectory (labelled ``empty map'') in the input trajectory. The ``empty map'' approach starts a new map during the test trajectory. This prevents loop closures between the input and test trajectory for the empty map approach. If the empty map performance is close to the full map, we expect a trajectory is not sensitive to the selection of map points for the input map.
}
                    \label{fig:base-acc-euroc}
        \end{table}

Unlike the KITTI dataset, we find that the relative pose error metric is not significantly affected by the selected map points. Therefore, this paper chooses to use absolute pose error for this dataset. The trajectories with the largest difference in trajectory errors in Table \ref{fig:base-acc-euroc} are MH05, V102, V103 and V202, and this paper consequently uses these trajectories to evaluate map point selection approaches. Due to the different properties of these trajectories, it is expected that good performance for map point selection will emphasise different aspects of the SLAM implementation than the KITTI trajectories.
        
\section{Experimental Results} \label{sec:experiment}
This section evaluates the proposed approaches against potential alternatives and relevant baselines from Subsection \ref{subsec:baselines} using the experimental setup described in Section \ref{sec:setup}. This section is divided into the following subsections: First, we list all the map point selection approaches compared in Subsection \ref{subsec:selectapproaches}. Secondly, the practicality of different approaches is evaluated by comparing sample execution times in Subsection \ref{subsec:timing}. Thirdly, the trajectory accuracy and loop-closure performance on the KITTI dataset are evaluated in Subsection \ref{subsec:res-kitti}. Evaluation results for the EuRoC dataset is presented in Subsection \ref{subsec:res-euroc}. Lastly, we evaluate the proposed approaches online in Subsection \ref{subsec:online}.

\subsection{Tested Map Point Selection Approaches}\label{subsec:selectapproaches}
    This subsection provides an overview of the different map point selection approaches we choose to evaluate. This paper proposes a series of different utility functions that can be optimised using the greedy algorithms from Subsection \ref{subsec:framework}. 
    
    We evaluate the proposed proposed utilities based on SLAM entropy, the localisation approximation and the visual odometry approximation as described in Section \ref{sec:traj}, and optimise these approaches using the lazy greedy algorithm. We denote these approaches by \emph{SLAM}, \emph{odometry} and \emph{localisation}, respectively. We also test the potential speedup of using the stochastic greedy algorithm for the baseline SLAM, odometry and localisation utilities with $\epsilon = 0.05$. We denote these approaches as \emph{SLAM-stoch}, \emph{odom-stoch} and \emph{local-stoch}.
    
    For the KITTI trajectories that feature large-scale loop closures, we also test the impact of extending the odometry utility with additional loop\hyp{}closure information as proposed in Section \ref{sec:loop}. We use the combined coverage approach denoted by \emph{odom+cover} with $b_{\text{cover}} = 300$ and optimise the resulting algorithm with the lazy greedy algorithm. This combined approach is aimed at improving trajectory estimation accuracy for datasets with large-scale loop closures. Due to the lack of large scale loop-closures in the EuRoC dataset, we do not evaluate the performance of the combined approach on the EuRoC dataset. 
    
    All approaches are modified to use the simple heuristic of including all the map points observed in the last keyframe. This heuristic ensures that the ORB-SLAM 2 implementation can continue tracking during the test trajectory after the map has been reduced. The cost of this heuristic is often negligible compared to the rest of the map. Applying the same heuristic across the different methods ensures a fair evaluation of the respective algorithms. 
    
    The coverage\hyp{}based integer program proposed by \citet{dymczyk2015gist} is also tested as an alternative technique. This approach was previously applied to localisation problems and has not yet been applied to SLAM. Our implementation uses the commercial Gurobi \cite{gurobi} optimiser to solve the resulting integer programming problem. To adapt the formulation to the SLAM problem, an additional constraint is added to include all map points in the last keyframe. This approach was tested with a coverage value, or $b$, of 100 and adjustable weight, or $\lambda$, of 25. This approach with the associated chosen parameters is denoted as \emph{ip100}. We provide a brief overview of this model and the relevant parameters in \ref{sec:ip-model}. 
    
\subsection{Timing Results for KITTI and EuRoC Datasets}\label{subsec:timing}

\begin{figure}[htbp]
        \centering
            \input{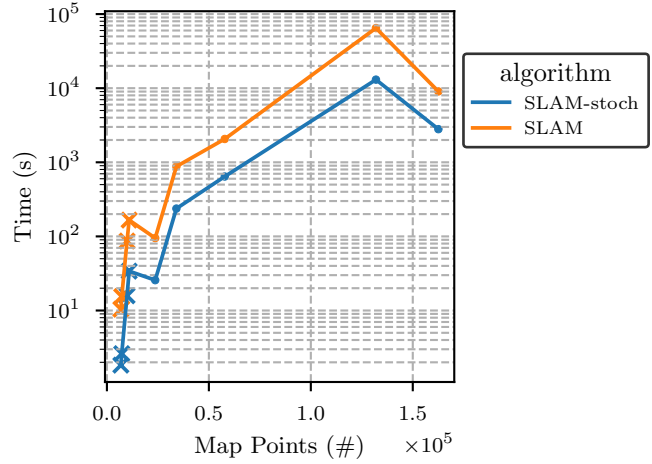}
            \caption{Execution times for the SLAM for the SLAM approach using either the lazy greedy algorithm (labelled ``SLAM'') or the stochastic greed algorithm (labelled ``SLAM-stoch''). Execution times are for selecting 15\% of the total number of map points for several trajectories in both the EuRoC and KITTI datasets. A cross denotes a trajectory in the EuRoC dataset and a dot denotes a trajectory in the KITTI dataset.
        }
    \label{fig:timing-slam}
\end{figure}

\begin{figure}[htbp]
        \centering
            \input{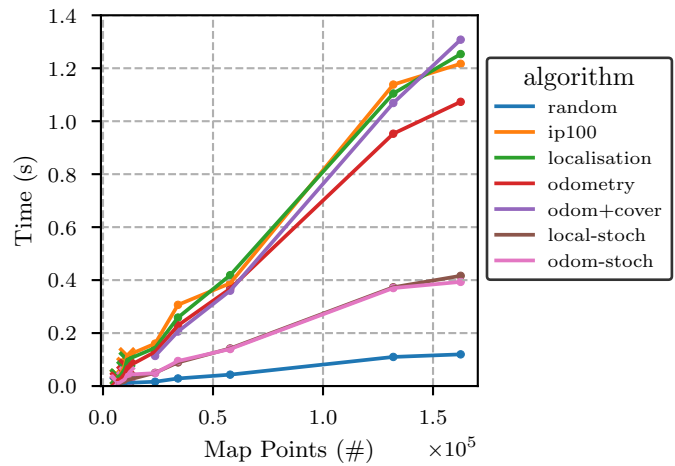}
            \caption{Execution times for online map point selection approaches versus the total number of map points. Listed execution times are for selecting 15\% of the total number of map points for several trajectories in both the EuRoC and KITTI datasets. A cross denotes a trajectory in the EuRoC dataset and a dot denotes a trajectory in the KITTI dataset.
        }
    \label{fig:timing-graph}
\end{figure}

The scalability of different approaches is measured by evaluating the average execution time required to reduce each input map to 15\% of its original size. For every selected trajectory in the KITTI and EuRoC dataset, we generate five different input maps using the initial input trajectory and apply each map point selection approach to this set of input maps. The results are generated on a personal computer that uses Linux, a Ryzen R9-3900x desktop CPU (with a PassMark score of 32200) and 16 GB RAM. The results for the SLAM approach using the lazy and stochastic greedy algorithm are shown in Figure \ref{fig:timing-slam}. The results for approaches suitable for online use, that is approaches except the two SLAM approaches, are shown in Figure \ref{fig:timing-graph}. These results are also reported in \ref{sec:timing-appendix} in tabular format. 

The timing results from Figure \ref{fig:timing-slam} show that the SLAM approach is both expensive and scales poorly to larger settings. The execution time for shorter trajectories often takes multiple minutes for shorter KITTI trajectories and increases to multiple hours for the KITTI 00 trajectory. Even for the smaller EuRoC datasets, map point selection can take more than a minute. These results show that the SLAM approach is only suited for very small environments or as an offline baseline. Despite the superior asymptotic complexity of the SLAM-stoch approach to optimise the SLAM utility function, using stochastic greedy instead of lazy greedy improved execution times by roughly a factor of three for the tested scenarios.

Subsection \ref{subsec:entropy} showed that the complexity of the SLAM utility is not only a function of the map points and keyframes, but also the density of the information matrix. This dependency results in longer execution times for the KITTI 00 trajectory (which has 130 000 map points), despite the resulting input map containing fewer map points than the KITTI 02 trajectory (which has 160 000 points).

As shown in Figure \ref{fig:timing-graph}, execution times for many of the online approaches (that is, approaches except for the SLAM approach) remained similar, increasing roughly linearly with increasing map size to about ~1.2 seconds. This roughly linear increase is despite significant differences in the worst-case asymptotic complexity. The ip100 approach has a non-polynomial complexity, and we expect that for large enough problems, this approach would be outperformed by approaches with polynomial-time complexity. However, for the tested scenarios, this only required 1.2 seconds for the largest map in the tested datasets. Note that integer programming solvers employ heuristic steps when optimising over the search space, which can lead to significant speed-ups in practice. 

Execution times for the odometry and localisation approaches are comparable to the ip100 approach. The odometry approach was slightly faster, taking 1.07 instead of 1.25 for the KITTI 02 dataset. These execution times scale roughly linearly in this range of map sizes, despite the $O(n^2)$ worst-case complexity for using the lazy greedy algorithm. While we found that the odometry and localisation approaches are already very scalable, we also tested the potential speed-up using stochastic greedy instead. Despite the superior asymptotic complexity of $O(n)$, using stochastic greedy resulted in roughly a constant factor improvement of 3. Given that the odometry and localisation approaches are already suitable for online use even when using lazy greedy, we continued to use lazy greedy for deterministic selection and stronger performance guarantees. When using these approaches in larger environments with limited computing, it might be necessary to use stochastic greedy and adjust the degree of approximation, $\epsilon$, based on the processing time available. 

Lastly, when using the combined odom+cover approach from Section \ref{sec:loop}, the additional term coverage term resulted in a small performance penalty compared to the odometry approach for maps exceeding 100 000 map points. This approach required a maximum of 1.3 seconds for the KITTI 02 dataset, compared to the 1.07 seconds for the odometry approach. For smaller datasets, the performance penalty was less than 30 ms. 

All approaches, except those based on the SLAM utility, have execution times in the order of a second or less. This is often far shorter than the time required to perform SLAM estimation or traverse environments of this scale. These relatively short execution times show that these methods are viable for online use. The following sections evaluate the resulting trajectory accuracy and, therefore, the quality of the maps produced by different approaches.

\subsection{Trajectory Accuracy for the KITTI Dataset }\label{subsec:res-kitti}

\begin{figure}[tbp]
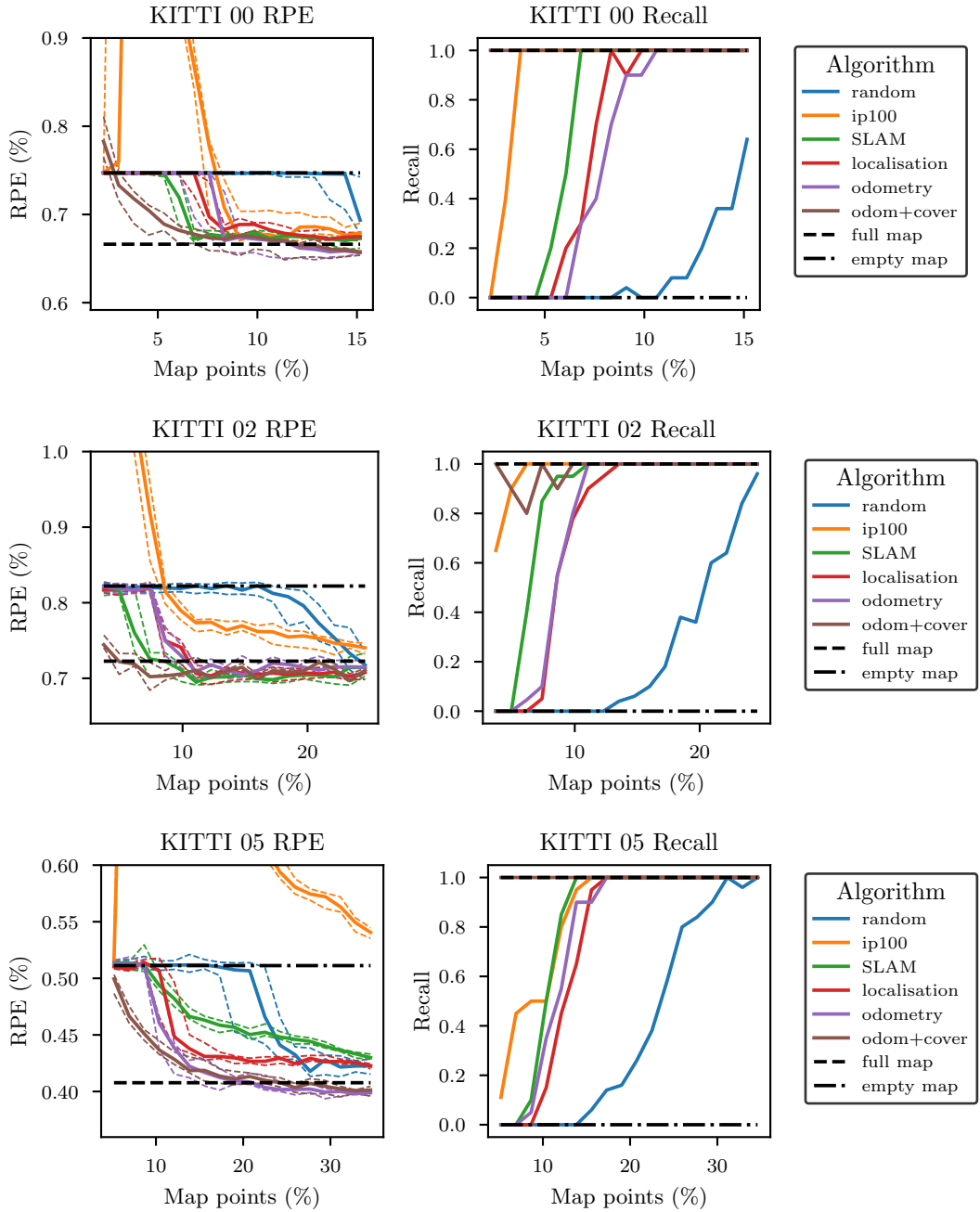

        \centering
            \input{fig_all_kitti00_4300_rpe_recall.pgf}
            \input{fig_all_kitti02_4100_rpe_recall.pgf}
            \input{fig_all_kitti05_2200_rpe_recall.pgf}
            \caption{RPE and recall for the KITTI 00, 02 and 05 trajectories as a function the map point budget. 
        Proposed approaches are compared against the empty map, full map and random baselines and alternative integer program approach labelled ip100 \cite{dymczyk2015gist}.
        }
            \label{fig:kitti-ent-map}
\end{figure}

The previous section evaluated the execution times of different approaches to map point selection. This section evaluates the SLAM trajectory accuracy and loop\hyp{}closure performance for the KITTI dataset using the proposed modified version of ORB-SLAM 2 and the testing setup described in Section \ref{sec:setup}: for each trajectory and map point budget, a selection approach is tested ten times, except for random, which is tested with 50 samples. The median of the relative pose error as well as the first and third quantile are shown in Figure \ref{fig:kitti-ent-map} and Figure \ref{fig:kitti-ent-map2}. These figures also include the average recall for the respective map point selection approaches. The ``empty map'' and ``full map'' baselines are not tied to a specific budget, but the performance of these baselines are plotted as a reference for the relative performance of map point selection approaches at a specified budget.

\begin{figure}[tbp]
    \centering
        \input{fig_all_kitti06_750_rpe_recall.pgf}
        \input{fig_all_kitti07_900_rpe_recall.pgf}
        \caption{RPE and recall for the KITTI 06 and 07 trajectories as a function of the map point budget. 
        Proposed approaches are compared against the empty map, full map and random baselines and alternative integer program approach labelled ip100 \cite{dymczyk2015gist}.
        }
        \label{fig:kitti-ent-map2}
\end{figure}

As shown in Figures \ref{fig:kitti-ent-map} and \ref{fig:kitti-ent-map2}, the random selection approach often also had trajectory accuracy close to the full map when provided with enough map points to obtain good recall in a given trajectory. Conversely, the random selection approach had a trajectory accuracy comparable to the empty map approach no loop\hyp{}closures were found. The map point budget for achieving good recall for the random approach varies between 15\% for the KITTI 00 trajectory and 40\% for the KITTI 07 trajectory. This comparable trajectory accuracy results from ORB-SLAM 2 only updating trajectory estimates for the full map after loop closures. As a result, when random selection fails to achieve good recall at a given budget, it performs poorly. 

The trajectory accuracy of the random selection approach indicates the relative difficulty of selecting map points for that trajectory. The loop closure in the KITTI 07 trajectory has minimal overlap between frames in the input portion of the trajectory and when revisiting the area during testing. This contributes to the increased difficulty of achieving good recall for this trajectory.

The ip100 approach offers good recall performance compared to many other approaches, achieving higher recall than many alternatives at the same budgets in the recall graphs for KITTI 00 through 07. This approach was proposed to select map points for localisation and continue to provide good recall on this dataset. Despite its good recall performance, the ip100 approach performed poorly regarding trajectory accuracy, often performing worse than random selection or even had a worse trajectory accuracy than the empty map approach. The ip100 approach also fails to deliver competitive trajectory accuracy at higher budgets and is often outperformed by random selection at higher budgets, as seen in the KITTI 02, 05, 06 and 07 trajectories. This shows that good recall performance does not necessarily lead to good SLAM trajectory accuracy.

The baseline SLAM approach achieved trajectory accuracy comparable to the full map in the KITTI 00 and 02 trajectories for map point budgets larger than 10\% and consistently outperforming random selection. For the KITTI 05 trajectory, the SLAM approach outperformed random selection for the range between 10\% and 25\% map points, but random selection achieved higher trajectory accuracies for larger maps. Results from the KITTI 06 trajectory showed SLAM approach generated a map comparable to the full map for most of the tested budgets. For the KITTI 07 trajectory, the SLAM approach was comparable to the random approach. This indicates that the SLAM approach did not accurately capture the utility of map points for the KITTI 05 and 07 trajectories.

The SLAM approach makes two important assumptions: First, it assumes that the information gain is a good approximation of the information gain of the posterior distribution that includes future measurements. This assumption is made out of convenience and will not hold for all trajectories. The second assumption is that the chosen model of the posterior is an accurate model for the impact of map points on trajectory accuracy. Various practical effects, for example, linearisation errors, missed loop\hyp{}closures or outliers, result in the baseline SLAM approach not accurately modelling the utility of map points in all cases. As a result, the baseline SLAM approach will not necessarily outperform alternatives in practice. 

Results from the localisation approach show that this approach needed more map points than the SLAM approach to achieve comparable trajectory accuracy in the KITTI 00 and 02 trajectories. The localisation approach offered better trajectory accuracy than the SLAM approach in the KITTI 05 trajectory and slightly better than the trajectory accuracy of the full map in the KITTI 06 trajectory. Similar to the SLAM approach, the localisation does worse for the KITTI 07 trajectory, with performance comparable to the random approach. For these trajectories, the localisation approach is a practical approximation of the SLAM approach, often offering comparable performance at reduced computational complexity.  

The odometry approach offers comparable trajectory accuracy to the SLAM approach at higher map point budgets in the KITTI 00 trajectory but is outperformed by the SLAM and localisation approaches at lower map point budgets. The worse trajectory accuracy of the odometry approach for the KITTI 00 trajectory coincides with worse recall performance at the tested map point budgets. For the KITTI 02 trajectory, the odometry approach requires larger maps than the SLAM approach to achieve equivalent trajectory accuracy. The odometry approach consistently outperformed the localisation, SLAM and random approaches in the KITTI 05 trajectory and obtained performance equivalent to the full map for a large range of map budgets. The odometry approach also outperforms the previously mentioned approaches for the KITTI 06 dataset, where it does notably better than the full map baseline at lower budgets. This approach performs similar to the SLAM, localisation and random approaches for the KITTI 07 dataset. The odometry utility seems to capture the utility of map points for trajectory estimation better than the localisation approach based on performance for the KITTI 00 and 05 trajectory. 

As shown in Figures \ref{fig:kitti-ent-map} and \ref{fig:kitti-ent-map2}, we found that the odometry+cover improve the loop\hyp{}closure performance of the odometry approach. This results in improved trajectory accuracy compared to the odometry approach for smaller map point budgets where loop\hyp{}closure performance limited performance. Note that since the KITTI 06 trajectory already achieved good recall performance over the tested range of map point budgets, it does not meaningfully benefit from this combined approach. These combined approaches outperform alternatives both in terms of trajectory accuracy and recall for the KITTI trajectories. For larger map point budgets, these approaches obtain comparable trajectory accuracy to the odometry approach. These combined approaches are the only approaches that could consistently close loops for the KITTI 07 trajectory at a map point budgets as low as 15\%. This shows that not only do the different modules of a SLAM implementation, such as the loop\hyp{}closure heuristics, impact performance but additional information, such as the potential location of loop-closures, can improve performance can lead to improved map point selection. 

\subsection{Trajectory Accuracy for the EuRoC Dataset}\label{subsec:res-euroc}
Subsection \ref{subsec:res-kitti} tested different map point selection approaches on large-scale SLAM problems in the KITTI dataset. This subsection evaluates performance on the EuROC dataset that features more dynamic motion and numerous shorter-term loop closures in room-sized environments. 

 As discussed in Subsection \ref{subsec:base-euroc}, the relative pose error was not significantly affected by map point selection, and as a result, we use absolute pose errors for this dataset.

Due to the lack of explicit large-scale loop closures, we do not evaluate recall performance or test approaches that use additional loop\hyp{}closure information, such as odom+cover on this dataset.

\begin{figure}[tbp]
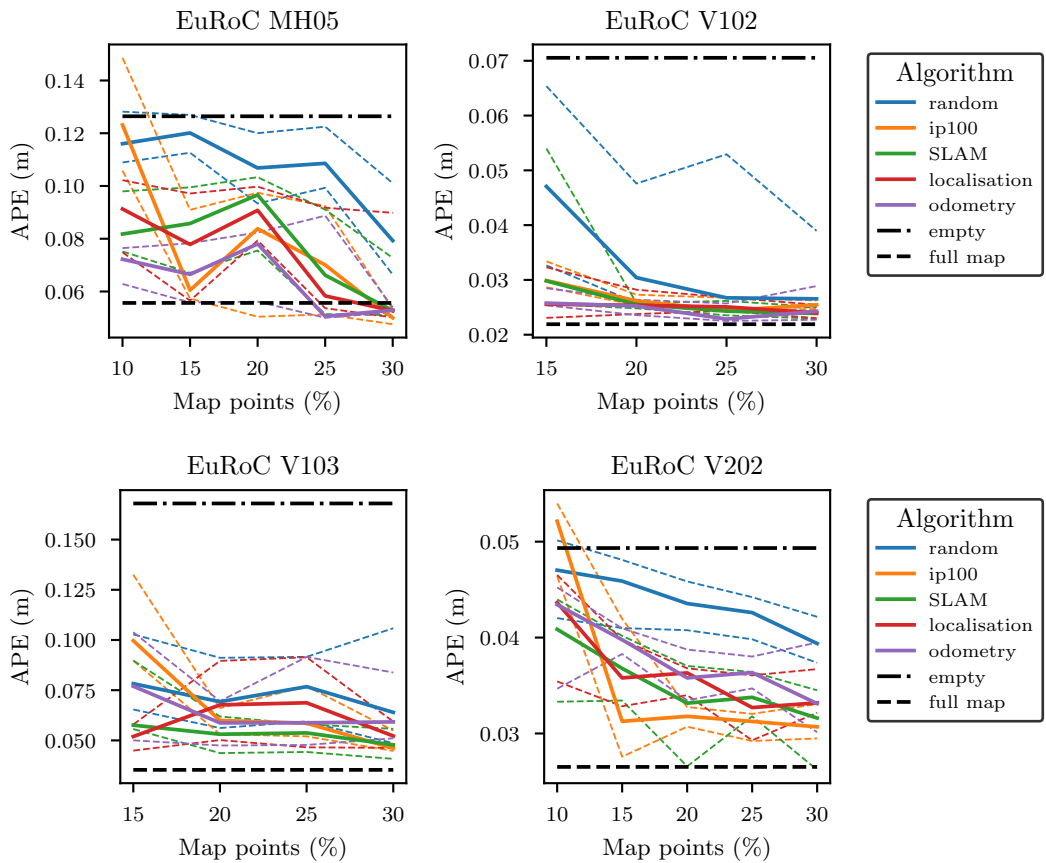

    \centering
    \input{fig_euroc-0.pgf}
    \input{fig_euroc-1.pgf}
    \caption{Evaluation of map point selection approaches on the chosen EuRoC trajectories. The median absolute pose error (APE) is shown over the tested range of map point budgets. The EuRoC dataset does not feature similar large-scale loop closures as the KITTI dataset; therefore, we do not report recall for this dataset. 
    }
    \label{fig:euroc-ape}
\end{figure}

Similar to the KITTI dataset, we evaluate different map point selection approaches by first generating different input maps using an initial input portion of the trajectory. The different map point selection approaches are applied to these input maps and tested with the modified ORB-SLAM on the remaining test portion of the trajectory. We report the APE's median, first and third quantile for the ten samples for each of the chosen EuRoC trajectories in Figure \ref{fig:euroc-ape} (50 samples for random). 

Experimental results from Figure \ref{fig:euroc-ape} show that most of the tested map point selection approaches outperformed random selection over the tested range of map point budgets. However, the trajectory accuracy of the different approaches, that is, ip100, SLAM, localisation and odometry, is often not significant compared to the variation in trajectory accuracy. The ip100 approach is more competitive in this dataset than in the KITTI dataset.

The trajectories in the EuRoC dataset lack large-scale loops and traverse the same environment multiple times. This reduces the importance of selecting high-quality map points for SLAM estimation as future observations can allow accurate reconstruction of the map in areas affected by selection. The diminished value of bundle adjustment for ORB-SLAM 2 in the EuRoC dataset is also supported by the fact that ORB-SLAM 3 in \citet{ORBSLAM3} no longer uses bundle-adjustment for this dataset and instead only used pose-graph estimation after the first few keyframes.

For the EuRoC trajectories, particularly the V103 and MH05 trajectory in the 10\% to 20\% range, trajectory accuracy improves in a less consistent fashion with the number of selected map points than previously observed for the KITTI dataset. We suspect that due to the lack of large-scale loop closures, the effect of other components of the SLAM implementation (such as tracking or local mapping) more significantly impacts the trajectory accuracy. These effects are unmodelled by the utility function and trajectory accuracy of the overall SLAM implementation is not necessarily be monotone with increasing map points. If, for example, a very limited number of map points was selected in the region, the tracking module will insert more keyframes due to the limited number of tracked map points, which could potentially result in improved accuracy. Similarly, selecting map points influences the co-visibility graph used for local SLAM, affecting which keyframes are included in the local SLAM  estimation. 

Any of the tested map point selection approaches are capable of reducing the map for the EuRoC dataset by 30\% of the map resulting in only a decrease in trajectory accuracy of 2 cm in the tested scenarios. Considering the accuracy and execution times, all the approaches except SLAM perform similarly for the EuRoC dataset. 

\subsection{Online Selection on the KITTI Dataset}\label{subsec:online}

The tests presented thus far are offline tests using the modified ORB-SLAM 2 algorithm to evaluate different map point selection approaches. These modifications were made to allow the evaluation of offline techniques and compare different selection approaches using the same set of maps. This subsection instead evaluates the ORB-SLAM 2 algorithm with map point selection online without the modifications of Subsection \ref{subsec:orb-mod}. 

As recommended in Subsection \ref{subsec:framework},  map point selection is implemented in parallel to the SLAM implementation. We start a new thread from local mapping with the map point exceeds a specified budget, and we are not currently busy with map point selection. Since new map points are added while the selection of map points is calculated in parallel, we need to compensate for these future map points. As a result, we need to reduce the total number of map points to a level below the budget. For simplicity, we reduce the total number of map points to 1000 map points below the map point budget.   

\begin{table}[htbp]
    \centering
    \begin{tabular}{l|ccc|ccc}
\hline
KITTI 00 & \multicolumn{3}{c|}{RPE (\%)} &  \multicolumn{3}{c}{APE (m)} \\
map points & 10k  & 20k  & 30k  & 10k & 20k & 30k \\
\hline
full map    & 0.70 & 0.70 & 0.70 	& 1.3 & 1.3 & 1.3 \\
ip100       & 0.82 & 0.77 & 0.73 	& 1.8 & 1.6 & 1.5 \\
local       & 0.76 & 0.69 & 0.70 	& 2.0 & 1.3 & 1.3 \\
odom        & 0.76 & 0.69 & 0.69 	& 2.3 & 1.3 & 1.3 \\
odom+cover  & 0.71 & 0.71 & 0.69 	& 1.3 & 1.3 & 1.2 \\
lt-odom     & 0.72 & 0.70 & 0.68 	& 1.3 & 1.3 & 1.2 \\
\hline
KITTI 02 & \multicolumn{3}{c|}{RPE (\%)} &  \multicolumn{3}{c}{APE (m)} \\
map points & 10k  & 20k  & 30k  & 10k & 20k & 30k \\
\hline
full map    & 0.76 & 0.76 & 0.76 	& 5.7 & 5.7 & 5.7 \\
ip100       & 1.02 & 0.89 & 0.86 	& 6.1 & 6.9 & 7.3 \\
local       & 0.86 & 0.80 & 0.83 	& 11.0 & 7.0 & 10.1 \\
odom        & 0.83 & 0.82 & 0.79 	& 8.5 & 7.9 & 6.9 \\
odom+cover  & 0.84 & 0.81 & 0.78 	& 5.7 & 5.4 & 6.0 \\
lt-odom    & 0.84 & 0.80 & 0.80 	& 6.9 & 6.0 & 6.0 \\
\hline
KITTI 05 & \multicolumn{3}{c|}{RPE (\%)} &  \multicolumn{3}{c}{APE (m)} \\
map points & 10k  & 20k  & 30k  & 10k & 20k & 30k \\
\hline
full map    & 0.40 & 0.40 & 0.40 	& 0.8 & 0.8 & 0.8 \\
ip100       & 0.46 & 0.44 & 0.42 	& 1.0 & 1.0 & 0.9 \\
local       & 0.47 & 0.40 & 0.39 	& 1.0 & 0.8 & 0.8 \\
odom        & 0.52 & 0.38 & 0.39 	& 2.0 & 0.7 & 0.7 \\
odom+cover  & 0.38 & 0.40 & 0.39 	& 0.7 & 0.8 & 0.7 \\
lt-odom   & 0.37 & 0.39 & 0.38 	& 0.7 & 0.7 & 0.7 \\
\hline
KITTI 06 & \multicolumn{3}{c|}{RPE (\%)} &  \multicolumn{3}{c}{APE (m)} \\
map points & 10k  & 20k  & 30k  & 10k & 20k & 30k \\
\hline
full map    & 0.51 & 0.51 & 0.51 	& 0.8 & 0.8 & 0.8 \\
ip100       & 0.54 & 0.53 & 0.54 	& 0.9 & 0.8 & 0.8 \\
local       & 0.51 & 0.54 & 0.51 	& 0.7 & 0.8 & 0.9 \\
odom        & 0.49 & 0.50 & 0.48 	& 0.9 & 0.7 & 0.7 \\
odom+cover  & 0.47 & 0.51 & 0.49 	& 0.6 & 0.8 & 0.7 \\
lt-odom    & 0.47 & 0.48 & 0.47 	& 0.7 & 0.8 & 0.7 \\
    \end{tabular}
    \caption{RPE and APE Errors for KITTI 00, 02, 05 and 06 trajectories using online selection. }
    \label{table:online-select}
\end{table}

\begin{table}[htbp]
    \centering
    \begin{tabular}{l|ccc|ccc}

\hline
KITTI 07 & \multicolumn{3}{c|}{RPE (\%)} &  \multicolumn{3}{c}{APE (m)} \\
map points & 10k  & 20k  & 30k  & 10k & 20k & 30k \\
\hline
full map    & 0.50 & 0.50 & 0.50 	& 0.5 & 0.5 & 0.5 \\
ip100       & 0.50 & 0.50 & 0.49 	& 0.6 & 0.5 & 0.5 \\
local       & 0.84 & 0.51 & 0.51 	& 1.4 & 0.5 & 0.6 \\
odom        & 0.68 & 0.49 & 0.50 	& 0.5 & 0.5 & 0.6 \\
odom+cover  & 0.50 & 0.49 & 0.51 	& 0.6 & 0.5 & 0.4 \\
lt-odom     & 0.50 & 0.49 & 0.51 	& 0.5 & 0.5 & 0.5 \\
    \end{tabular}
    \caption{RPE and APE Errors for KITTI 07 trajectory using online selection. }
    \label{table:online-select2}
\end{table}

We evaluate the online ip100, localisation, odometry and the combined odom+cover apporaches on the selected KITTI trajectories. For the purposes of demonstration, we also include a test where we halved the loop-closure threshold of ORB-SLAM 2 from 40 map points to 20 map points and label this approach ``lt-odom''. For each of the KITTI trajectories, we test map point budgets of 10k, 20k and 30k. We repeat this five times and report the median RPE and APE in Tables  \ref{table:online-select} and \ref{table:online-select2}.

Tables \ref{table:online-select} and \ref{table:online-select2} shows that the proposed utility functions allow limiting the size of the map while maintaining trajectory accuracy comparable to the full map without selection. When the available map point budget starts to affect loop closures - for example, when choosing a map point budget of 10k for the KITTI 05 or 07 trajectory, the trajectory accuracy of the odometry and localisation approaches drops significantly. In these cases, the combined odom+cover approach results in improved performance. For map point budgets where loops can be consistently closed, the odometry technique provides trajectory accuracy comparable to the full map.  

The discrepancy in trajectory accuracy between the techniques is also less pronounced than for the offline results. For the offline results, we increased the number of bundle adjustment iterations, allowing the full SLAM estimation to estimate the trajectory accurately. In these online tests, a heavier emphasis is on the built-in heuristics (specifically the pose graph). This leads to comparatively worse results than for the offline case. ORB-SLAM 2 has been designed for visual SLAM without selection in mind. Some of the heuristic mechanisms in ORB-SLAM that work well without selection cause comparatively bad performance with selection, even if a good selection is made, as shown by the difference between the offline and online results. 

The loop-closure threshold demonstrates the importance of properly adjusted heuristics when using maps alongside map point selection approaches. At these specific map point budgets, this was sufficient to allow the detection of the relevant loop-closures and improve performance for the odometry approach. 

In summary, these results demonstrate that map point selection allows reducing the number of map points online, often with a minimal impact on trajectory accuracy - provided an appropriate map point budget is chosen. Additional work is still required to optimise ORB-SLAM 2 for the presence of map point selection. We expect that for an optimised implementation of ORB-SLAM 2 for map point selection, it will be beneficial to revisit heuristic choices related to the loop-closure detection, keyframe insertion, and the co-visibility graph for working on the sparse maps resulting from map point selection. 

\section{Conclusion}
\label{sec:conclusion} 
Existing sparse visual SLAM implementations do not limit the storage space required to store the map and associated visual feature information. This is undesirable for large-scale applications when dealing with resource-constrained platforms. This paper proposes the use of map point selection algorithms to limit the map's total number of map points and the associated feature information. There are several existing approaches for map point selection for large-scale \emph{localisation}. This paper instead investigates the more difficult and novel problem of selecting map points for \emph{SLAM}.

The main contribution of this paper is the development of novel information\hyp{}theoretic approaches for map point selection. These approaches use existing greedy techniques to optimise novel information-theoretic utility functions. The first utility function is an information\hyp{}theoretic utility using the full SLAM posterior. This function has a high asymptotic computational complexity even when optimisation is done with greedy algorithms. While this approach is not appropriate for large-scale problems, it serves as an offline baseline for other more scalable approaches. This paper also proposes two approximate utility functions that simplify the SLAM posterior based on an odometry and localisation approximation. These approximate utility functions significantly improve the asymptotic computational complexity compared to the baseline SLAM function, reducing execution times by multiple orders of magnitude on the tested datasets. These approximate approaches reduce kilometre scale maps in seconds instead of hours while offering similar trajectory accuracy to the offline SLAM approach. We found that the odometry approach performed marginally better than the localisation approach for the KITTI dataset. However, either approach provides competitive map point selection performance in various scenarios and map point budgets. These approaches require no parameters to be adjusted for specific datasets and are easy to implement.

The second contribution of this paper was showing that if additional information is available regarding the location of future loop closures, incorporating this information into the map point utility yields results in maps that outperform alternatives. The improved performance of this combined approach demonstrates the value of additional information for map point selection. However, this approach assumes that information about the location of future loop closures is available, which is not typically the case for existing SLAM implementations. Due to the improved performance that this information enables, these results highlight the value of future research to make this information available. A potential option for predicting the location of future loop closures is to use the robot's planned trajectory. Another option is to use information about the environment topology; for example, identifying scenes as road intersections could be used to increase the likelihood of a loop closure at that location for vehicle applications.

A limitation of our current software implementation is that it is not optimised for memory-constrained applications. The software implementation was designed to test a wide array of utility functions using the proposed testing procedure. A significant amount of memory is dedicated to loading shared libraries and reconstructing relevant data available to a SLAM algorithm from storage. An optimised implementation and integration alongside a SLAM implementation of the approximate information-theoretic approaches is the subject of future work. A challenge with this direction is that SLAM algorithms are not necessarily designed to be memory efficient. For example, ORB-SLAM 2 does not free memory related to deleted map points.

This paper developed scalable map point selection approaches that are an important step towards addressing the map point selection problem. Furthermore, the combined approach also shows that parts of the SLAM implementation typically not included in the back-end model of the SLAM problem, such as loop-closure detection, can play an essential role in the performance of map point selection approaches. We suspect that for robust solutions to the map point selection problem, it will be necessary to consider how map point selection affects the overall SLAM implementation and design SLAM algorithms with map point selection in mind.

\appendix

\section{Coverage Model}\label{sec:ip-model}
    This appendix provides a overview linear integer program proposed by \citet{dymczyk2015keep} for map point selection. This approach defines a minimisation problem over the vector $\mathbf{x}$ that encodes the selected set of map points. Given a budget of $n$ map points and desired number of map points visible in each frame $b$, this formulation minimises the weighted sum of two terms: a linear cost associated with the selected set of map points $\mathbf{q}$ and penalty term that penalises the objective proportion to the degree that the selected set of map points does not comply with the coverage requirement, or 
    \begin{equation}
    \begin{split}
        \min \mathbf{q}^T \mathbf{x}  &+ \lambda \mathbf{1}^T \boldsymbol{\zeta} \\
    \mathbf{A}\mathbf{x} &+ \boldsymbol{\zeta}  \geq b \mathbf{1} \\
        \sum\limits_{n=1}^{N}\mathbf{x}_i & = n,
    \end{split}
    \end{equation}
    where $\lambda$ is a design variable that controls the trade off between coverage, and the linear objective and $\cMat{A}$ is the matrix that encodes whether a map point is visible from the respective camera views. Each term $\mathbf{q}_i$ is assigned to be the difference between highest number of frames any map point is visible from and the frames the map point at index $i$ was visible from. 

\section{KITTI Run Variation} \label{sec:kitti-map-variation}

 The chosen trajectories for the KITTI dataset can also be used to demonstrate important details regarding our test procedure. Due to the stochastic nature of the ORB-SLAM 2 algorithm, maps generated by the algorithm differ between runs. If this variation is significant, it should be considered when evaluating map point selection approaches. This appendix shows the significant variation between different input maps. 
            \begin{figure}[h]
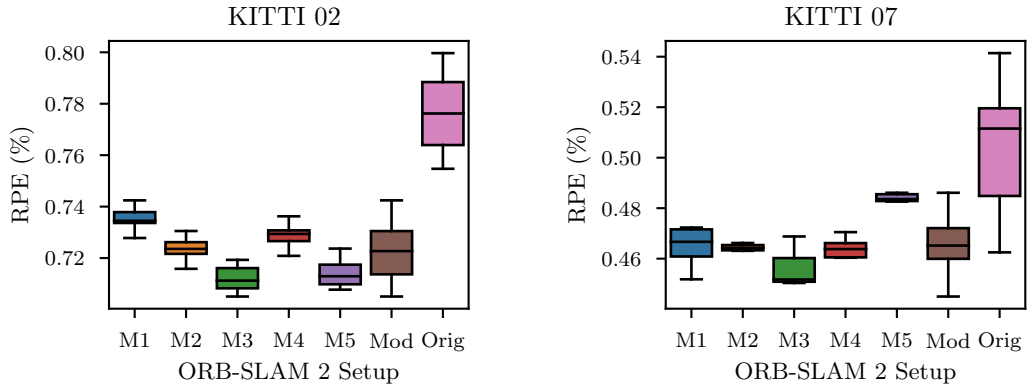

             \centering
             \begin{subfigure}[b]{0.45\textwidth}
                 \centering
                 \input{fig_RunVariation02-4100.pgf}
                 \label{fig:y equals x}
             \end{subfigure}
             \hfill
             \begin{subfigure}[b]{0.45\textwidth}
                 \centering
                 \input{fig_RunVariation07-900.pgf}
                 \label{fig:three sin x}
             \end{subfigure}
             \hfill
             \caption{
             Comparative analysis of the trajectory accuracy between various ORB-SLAM 2 setups for the KITTI 02 and 07 trajectories." These figures generate results from the five input maps labelled ``M1'' through ``M5'' using the modified ORB-SLAM 2 algorithm (described in Subsection \ref{subsec:orb-mod}) on the same trajectory. These maps are evaluated on the remaining test portion of the trajectory using RPE. ``Mod'' is the summary using samples from all five input maps used for evaluation during this work. For comparison, we include the variation of the original ORB-SLAM 2 algorithm (denoted ``Orig'') on the same trajectory.
            }
            \label{fig:run-variation}
        \end{figure}
    
   Using the test procedure described in Subsection \ref{subsec:testproc}, we compare the trajectory accuracy of different input maps generated on the same trajectory. To gather additional data, each of the five input maps is tested ten times and evaluate separately. Figure \ref{fig:run-variation} shows relative pose error for each estimated trajectory from this experiment for the KITTI 02 and 07 trajectories and a comparison to the original ORB-SLAM 2 algorithm. 

    Figure \ref{fig:run-variation} demonstrates that the relative pose errors vary significantly between the different input maps for the same trajectory. This shows the importance of evaluating map point selection approaches on similar maps. Since using a single map might not be representative, Subsection \ref{subsec:testproc} instead uses a set of five input map samples to test different map point selection approaches.

\section{Tables of Timing Results for KITTI and EuRoC}\label{sec:timing-appendix}
The average execution times are reported in Tables \ref{tab:timing} and \ref{tab:timing-euroc} for the KITTI and EuRoC datasets, respectively.

\begin{table}[htbp]
   \centering
    \begin{tabular}{l|rrrrr}
        \toprule
        
Algorithm  & 00   & 02   & 05   & 06    & 07    \\
\midrule
random              & 0.11 s    & 0.12 s    & 0.04 s    & 0.03 s    & 0.02 s    \\
ip100               & 1.14 s    & 1.22 s    & 0.39 s    & 0.31 s    & 0.16 s    \\
local               & 1.10 s    & 1.25 s    & 0.42 s    & 0.26 s    & 0.14 s    \\
odom                & 0.95 s    & 1.07 s    & 0.37 s    & 0.23 s    & 0.13 s    \\
odom+cover          & 1.07 s    & 1.31 s    & 0.36 s    & 0.21 s    & 0.11 s    \\
local-stoch         & 0.37 s    & 0.42 s    & 0.14 s    & 0.09 s    & 0.05 s    \\
odom-stoch          & 0.37 s    & 0.39 s    & 0.14 s    & 0.09 s    & 0.05 s    \\
SLAM                & 17 h 58 min& 2 h 30 min& 34 min 23 s& 14 min 37 s& 1 min 34 s\\
SLAM-stoch          & 3 h 37 min& 46 min 39 s& 10 min 42 s& 3 min 57 s& 25.7 s    \\

        \bottomrule
    \end{tabular}
    \caption{Execution times for different map point selection approaches on the KITTI trajectories. Maps are generated using the first input portion of the trajectories and reduced to 15\% map points using the respective map point selection approaches. The reported execution times are the average of ten runs of the approach. }
    \label{tab:timing}
\end{table}  

\begin{table}[htbp]
   \centering
    \begin{tabular}{l|rrrrr}
        \toprule
Algorithm           & MH05 & V102  & V103 & V202 \\
\midrule
random              & 0.01 s    & 0.01 s    & 0.01 s    & 0.01 s    \\
ip100               & 0.11 s    & 0.03 s    & 0.03 s    & 0.08 s    \\
SLAM                & 1 min 40 s& 6.12 s    & 9.02 s    & 51.2 s    \\
SLAM-stoch          & 34.0 s    & 1.83 s    & 2.62 s    & 15.8 s    \\
localisation        & 0.23 s    & 0.08 s    & 0.08 s    & 0.15 s    \\
odometry            & 0.32 s    & 0.12 s    & 0.12 s    & 0.22 s    \\
        \bottomrule
    \end{tabular}
    \caption{
    Execution times for different map point selection approaches on the EuRoC trajectories. Maps are generated using the first input portion of the trajectories and reduced to 15\% map points using the respective map point selection approaches. The reported execution times are the average of ten runs of the approach. }
    \label{tab:timing-euroc}
\end{table}

\clearpage
\section{APE Results}

\begin{figure}[htbp]
        \centering
            \input{fig_all_kitti00_4300_ape_recall.pgf}
            \input{fig_all_kitti02_4100_ape_recall.pgf}
            \caption{
            APE and recall for the KITTI 00 and 02 trajectories as a function the map point budget. Proposed approaches are compared against the empty map, full map and random baselines and alternative integer program approach labelled ip100 \cite{dymczyk2015gist}.
            }
            \label{fig:kitti-ape-map}
\end{figure}

\begin{figure}[htbp]
    \centering
        \input{fig_all_kitti05_2200_ape_recall.pgf}
        \input{fig_all_kitti06_750_ape_recall.pgf}
        \input{fig_all_kitti07_900_ape_recall.pgf}
        \caption{APE and recall for the KITTI 05, 06 and 07 trajectories as a function of the map point budget. Proposed approaches are compared against the empty map, full map and random baselines and alternative integer program approach labelled ip100 \cite{dymczyk2015gist}.
        }
        \label{fig:kitti-ape-map2}
\end{figure}
\clearpage





\bibliographystyle{elsarticle-num-names}

\bibliography{MapPointSelectionForVisualSLAM.bib}







\end{document}